\documentclass[journal]{IEEEtran}

\usepackage{bm}
\usepackage{amsthm}
\usepackage{amssymb}
\usepackage{amsmath}
\usepackage{mathtools}
\usepackage{algorithm2e}
\usepackage{setspace}
\usepackage{enumitem}
\usepackage{float}
\usepackage{balance}
\usepackage{color}
\usepackage[pdftex]{graphicx}
\usepackage[hidelinks]{hyperref}
\usepackage[font=footnotesize]{caption}
\usepackage{cite}

\newtheorem{prop}{Proposition}
\newtheorem{rem}{Remark}
\newtheorem{lem}{Lemma}

\DeclareMathSizes{10}{10}{6}{6}

\allowdisplaybreaks

\makeatletter
\newcounter{MyEqNum}
\newenvironment{mynumbering}{%
	
	\let\c@equation\c@MyEqNum 
	\let\p@equation\p@MyEqNum
	
}{}

\begin{document}
\title{Transferring Model Structure in Bayesian Transfer\\Learning for Gaussian Process Regression}

\author{Milan~Pape\v{z} and Anthony~Quinn%
	\thanks{M.\ Pape\v{z} is with the Institute of Information Theory and Automation (UTIA), Czech Academy of Sciences, Czech Republic, Prague 18208,
		Czech Republic (papez@utia.cas.cz).}%
	\thanks{A.\ Quinn is with UTIA (aquinn@utia.cas.cz), and also with the Department of Electronic and Electrical Engineering, Trinity College Dublin, the University of Dublin, Dublin D04 V1W8, Ireland (aquinn@tcd.ie).}%
	\thanks{\hspace{-8pt}The research has been supported by GA\v{C}R grant 18-15970S.}%
}

\maketitle

\begin{abstract}
Bayesian transfer learning (BTL) is defined in this paper as the task of conditioning a target probability distribution on a transferred source distribution. The target globally models the interaction between the source and target, and conditions on a probabilistic data predictor made available by an independent local source modeller. Fully probabilistic design is adopted to solve this optimal decision-making problem in the target. By successfully transferring higher moments of the source, the target can reject unreliable source knowledge (i.e.\ it achieves robust transfer). This dual-modeller framework means that the source's local processing of raw data into a transferred predictive distribution---with compressive possibilities---is enriched by (the possible expertise of) the local source model. In addition, the introduction of the global target modeller allows correlation between the source and target tasks---if known to the target---to be accounted for. Important consequences emerge. Firstly, the new scheme attains the performance of fully modelled (i.e.\ conventional) multitask learning schemes in (those rare) cases where target model misspecification is avoided. Secondly, and more importantly, the new dual-modeller framework is robust to the model misspecification that undermines conventional multitask learning. We thoroughly explore these issues in the key context of interacting Gaussian process regression tasks. Experimental evidence from both synthetic and real data settings validates our technical findings: that the proposed BTL framework enjoys robustness in transfer while also being robust to model misspecification.
\end{abstract}


\section{Introduction}
\IEEEPARstart{M}{ultitask} learning \cite{zhang2018survey} with Gaussian process (GP) regression tasks is of major concern in the statistical signal processing and machine learning communities. Many contemporary instances of this framework have been reported, including methods based on convolved GPs \cite{teh2005semiparametric,bonilla2008multi, nguyen2014collaborative,alvarez2011computationally}, convolutional GPs \cite{van2017convolutional}, product GPs \cite{wilson2012gaussian,nguyen2013efficient}, deep (nested) GPs \cite{damianou2013deep,kandemir2015asymmetric,boustati2019multitask}, and deep neural networks \cite{wilson2016deep}. All these methods adopt a single global model to describe the relationship among source and target tasks, involving a joint probability distribution of all latent and output-data processes, which is consistent with the foundational methodology of Bayesian learning \cite{bernardo1994bayesian}. The flexibility of multitask GP regression is evidenced by a wide range of applications, such as reinforcement learning \cite{lazaric2010bayesian}, Bayesian optimization \cite{swersky2013multi}, Earth observation \cite{camps2016survey}, face recognition \cite{zhang2010multi}, and clinical data analysis \cite{ghassemi2015multivariate}.

\begin{figure*}[t]
	\includegraphics{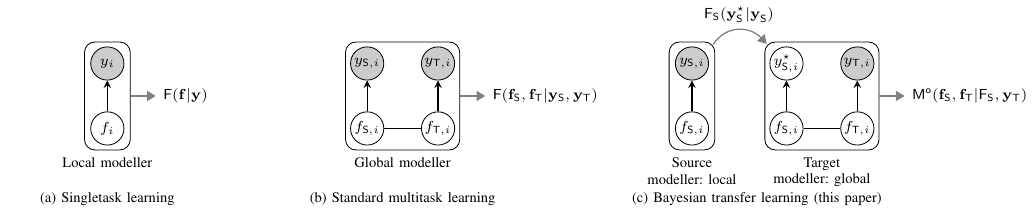}
	\caption{(a) \textit{Singletask learning}: an isolated local Bayesian modeller with complete knowledge of $\mathsf{F}(\mathbf{y},\mathbf{f})$. (b) \textit{Standard multitask learning}: an interacting global Bayesian modeller with complete knowledge of $\mathsf{F}(\mathbf{y}_{\mathsf{S}},\mathbf{y}_{\mathsf{T}},\mathbf{f}_{\mathsf{S}},\mathbf{f}_{\mathsf{T}})$. (c) \textit{Bayesian transfer learning}: an isolated local source Bayesian modeller with complete knowledge of $\mathsf{F}_{\mathsf{S}}(\mathbf{y}_{\mathsf{S}},\mathbf{f}_{\mathsf{S}})$ and an interacting global target Bayesian modeller with incomplete knowledge of $\mathsf{F}(\mathbf{y}^{\star}_{\mathsf{S}},\mathbf{y}^{}_{\mathsf{T}},\mathbf{f}^{}_{\mathsf{S}},\mathbf{f}_{\mathsf{T}},\mathsf{F}_{\mathsf{S}})$. The global target modeller does not have access to $\mathbf{y}^{}_{\mathsf{S}}$. It compensates for this missing information by utilizing the output-data predictive distribution, $\mathsf{F}_{\mathsf{S}}$, that is made available by the local source modeller. The global target modeller improves its performance based on $\mathsf{F}_{\mathsf{S}}$. The gray and white circles highlight know and unknown variables, respectively.}
	\label{graphical_model}
\end{figure*}

This paper is specifically interested in Bayesian transfer learning (BTL) \cite{torrey2010transfer,pan2010survey,weiss2016survey,lu2015transfer}. Without loss of generality, we define BTL as the transfer of knowledge from a source task to a target task, a definition which is widely extensible to multitask settings. Under the Bayesian framework, source and target modellers separately adopt distinct probability distributions to model local unknown quantities of interest, i.e.\ our approach involves \emph{two} stochastically independent modellers without requiring consistency to be established between them, whereas conventional multitask learning involves---as already stated---a single global modeller. These modellers each have exclusive access to their respective local data but not to each other's data. Therefore, the target cannot condition its model on the source data, the essential assumption of conventional global multitask learning. The target's model can be a local model, as in \cite{foley2018fully,papez2018dynamic,papez2019robust,jirsa2019knowledge,papez2019bayesian,papez2020bayesian}. Alternatively, it may itself extend its focus beyond its local target task and adopt a global model involving both source and target tasks, as in the standard multitask setting. This is the setup we adopt in this paper for the first time. To emphasize: the target modeller globally models the interactions between its own task and the source task, but is informed \emph{only} by its local target data. Meanwhile, the source modeller, adopting a local Bayesian model, is informed by its local (source) data, and then communicates its knowledge to the target only via a source probability distribution instantiated by its (source) sufficient statistics. Therefore, only the target takes care of interactions and only the source communicates source probabilistic knowledge. The key distinctions between global and local modelling of the source and target tasks are captured in Fig.\ \ref{graphical_model}, whose technical details will be explained fully in Section \ref{btl_for_global_modellers}. The target modeller has no prescriptive rule of probability calculus with which to condition on the source distribution in this BTL scenario (Fig.\ \ref{graphical_model}(c)). The reason for this is that the joint model of the source distribution and the target's quantities of interest is not available. The challenge for the target modeller then lies in deciding how to condition on the source probability distribution in this incompletely modelled case. Fully probabilistic design (FPD) \cite{karny2012axiomatisation,quinn2016fully}---the optimal Bayesian decision-making strategy based on the minimum cross-entropy principle \cite{shore1980axiomatic}---addresses this problem. Under this framework, the target modeller conditions on the transferred source knowledge by solving a constrained optimization problem.

The principal novelty of this paper, therefore, is that the target modeller is a global modeller (in sympathy with the aforementioned standard multitask learning) that conditions on a local source probability distribution. Recall, from Bayesian foundations, that this latter distribution is a function of both its local source data \emph{and} local source model. What is implied by this BTL framework is that the global target modeller independently adopts a different model from the source. This leads to important advances beyond the current available notions and techniques for BTL, all of which are explored in this paper:
\begin{itemize}[leftmargin=12pt]
	\item The introduction of the global target modeller facilitates learning about the correlation structure between the source and target tasks, extending our own previous BTL framework which involved only locally modelled tasks \cite{foley2018fully,papez2018dynamic,papez2019robust,papez2019bayesian,papez2020bayesian}. It also generalizes FPD-based approaches to pooling \cite{kracik2005merging,karny2006how}, as well as conventional (global) Bayesian multitask learning \cite{teh2005semiparametric,bonilla2008multi, nguyen2014collaborative,alvarez2011computationally,van2017convolutional,wilson2012gaussian,nguyen2013efficient,damianou2013deep,kandemir2015asymmetric,boustati2019multitask,wilson2016deep}. We demonstrate this formally, and also via extensive experimental evidence.
	\item We stand to benefit from the fact that the local source modeller is independent of the target and is therefore, potentially, a more informed---i.e.\ expressive or expert---modeller of its local data. This can provide a major performance dividend over standard multitask learning solutions that adopt only a single global (typically centralized) model.
	\item The source probability distribution is represented by its sufficient statistics, thereby achieving an optimal encoding (compression) of source knowledge extracted from its (often high-dimensional) data and (possibly complex) model structure. This can avoid transferred message overheads that arise in standard multitask learning, which relies on the transfer of unprocessed source data for its conditioning task (i.e.\ learning).
	\item Our previous FPD-based BTL schemes were unable to transfer higher moments of the source distribution, a resource that is necessary in achieving robust transfer (i.e.\ the rejection of imprecise source knowledge) \cite{azizi2016hierarchical,foley2018fully,papez2018dynamic,papez2019bayesian}. By careful design of the FPD-based constrained optimization problem in this paper, we formally solve this problem, obviating the informal adaptations in \cite{foley2018fully,papez2018dynamic}, and the computationally expensive augmentation strategies in \cite{papez2019robust,papez2020bayesian}.
\end{itemize}

The paper is organized as follows: Section~\ref{btl_for_global_modellers} defines the BTL problem, where the central aim is to transfer the source output-data predictor to the target modeller. The new concept of the global target modeller is introduced and the general solution based on the FPD framework is provided. Section~\ref{local_and_global_gpr_modellers} instantiates this general setting of BTL in the case of GP regression tasks. Section~\ref{synthetic_data_experiments} provides a thorough exploration of these advances via synthetic data experiments, focussing particularly on its ability to capture correlation structure, to achieve robust transfer and to benefit from the transfer of an expressive local model in the source. This will allow us to comment in an informed way on the performance limits of our approach. Section~\ref{real_data_experiments} validates our BTL scheme in a real-data context. Section~\ref{discussion} provides further discussion of the key properties of our method, including its source knowledge compression and computational complexity aspects. Section~\ref{conclusion} concludes the paper with suggestions for future work.

\section{Bayesian transfer learning from a local\\to global modeller}\label{btl_for_global_modellers}
Consider a (non-parametric) regression task where the aim is to infer unknown (latent) function values, $\mathbf{f}\equiv(f(x_{i}))_{i=1}^{n}$, based on known input data (regressors), $\mathbf{x}\equiv(x_{i}\in\mathbb{R}^{n_{x}})_{i=1}^{n}$, and known output data, $\mathbf{y}\equiv(y_{i}\in\mathbb{R})_{i=1}^{n}$, where each $y_{i}$ is an indirect (e.g.\ noisy) observation of $f(x_{i})\in\mathbb{R}$. A Bayesian modeller (Fig.\ \ref{graphical_model}(a)) addresses this problem by adopting a complete joint probability model, $\mathsf{F}(\mathbf{y},\mathbf{f}|\mathbf{x},\theta)$, and then computing the posterior distribution, $\mathsf{F}(\mathbf{f}|\mathbf{y},\mathbf{x},\theta)$, for some fixed and finite-dimensional parameter, $\theta\in\mathbb{R}^{n_{\theta}}$, given a realization of $\mathbf{y}$. This constitutes standard Bayesian conditioning in the context of the complete joint model\footnote{In the sequel, we will suppress the details of the conditioning where this is evident. Moreover, $\mathsf{F}$ and $\mathsf{M}$ denote known (fixed-form) and unknown (variational-form) distributions, respectively. We use ``$\equiv$'' to denote ``is defined to be equal to''.}, $\mathsf{F}(\mathbf{y},\mathbf{f}|\mathbf{x},\theta)$.

This paper addresses the problem of learning in a pair of GP regression tasks, called the source and target learning tasks, respectively. The source task is \emph{local}, being informed by local output data, $\mathbf{y}_{\mathsf{S}}$, available only to the source. Its local processing of knowledge, $\mathsf{F}_{\mathsf{S}}(\mathbf{y}_{\mathsf{S}},\mathbf{f}_{\mathsf{S}})$, and data, $\mathbf{y}_{\mathsf{S}}$, is to be exploited in order to improve learning at the target task. The standard approach is that a \emph{global} multitask modeller (Fig.\ \ref{graphical_model}(b)) elicits a joint model, $\mathsf{F}(\mathbf{y}^{}_{\mathsf{S}},\mathbf{y}^{}_{\mathsf{T}},\mathbf{f}^{}_{\mathsf{S}},\mathbf{f}^{}_{\mathsf{T}})$, capturing interactions between the source and target processes, and conditions on source and target output data, $\mathbf{y}_{\mathsf{S}}$ and $\mathbf{y}_{\mathsf{T}}$, respectively, to which it must have access \cite{teh2005semiparametric,bonilla2008multi, nguyen2014collaborative,alvarez2011computationally,van2017convolutional,wilson2012gaussian,nguyen2013efficient,damianou2013deep,kandemir2015asymmetric,boustati2019multitask,wilson2016deep}. This yields
\begin{equation}
	\mathsf{F}(\mathbf{f}^{}_{\mathsf{S}},\mathbf{f}^{}_{\mathsf{T}}|\mathbf{y}^{}_{\mathsf{S}},\mathbf{y}^{}_{\mathsf{T}})
	.
	\label{joint_model}
\end{equation}
This scheme requires the prior elicitation of the joint model of the two-task learning system. We refer to this as \emph{complete modelling}.

In this work, our aim is to relax these restrictive assumptions of the conventional multitask learning framework, as follows (Fig.\ \ref{graphical_model}(c)):
\begin{itemize}[leftmargin=16pt]
	\item[(i)] The target will not have direct access to the source data, $\mathbf{y}_{\mathsf{S}}$; and
	\item[(ii)] the target's joint model\footnote{We adopt the following notation: undecorated distributions, $\mathsf{F}$ and $\mathsf{M}$, are \emph{target} models, with $\mathsf{F}_{\mathsf{S}}$ denoting source distributions.}, $\mathsf{F}(\mathbf{y}^{}_{\mathsf{S}},\mathbf{y}^{}_{\mathsf{T}},\mathbf{f}^{}_{\mathsf{S}},\mathbf{f}^{}_{\mathsf{T}})$, is distinct from the source's local model, $\mathsf{F}_{\mathsf{S}}(\mathbf{y}_{\mathsf{S}},\mathbf{f}_{\mathsf{S}})$, and each is elicited independently.
\end{itemize}
Specifically, the goal of the present paper is to substitute the inaccessible source data, $\mathbf{y}_{\mathsf{S}}$, with appropriately specified probabilistic source knowledge, $\mathsf{F}_{\mathsf{S}}(\cdot)$. Technically, the inferential objective of the global target modeller is to compute $\mathsf{F}(\mathbf{f}_{\mathsf{S}},\mathbf{f}_{\mathsf{T}}|\mathsf{F}_{\mathsf{S}},\mathbf{y}_{\mathsf{T}})$, conditioned now on the transferred source distribution, $\mathsf{F}_{\mathsf{S}}$, but not on $\mathbf{y}_{\mathsf{S}}$. This conditioning on $\mathsf{F}_{\mathsf{S}}$ is incompletely specified because the target does not have a joint model of the type, $\mathsf{F}(\mathsf{F}_{\mathsf{S}}, \ldots)$, i.e.\ one that involves the target's hierarchical specification of the source's transferred distribution. Instead, the target's learning task becomes one of optimally choosing its knowledge-conditioned model, $\mathsf{M}(\ldots|\mathsf{F}_{\mathsf{S}},\mathbf{y}_{\mathsf{T}})$, in this \emph{incompletely modelled} case (necessarily, then, $\mathsf{M}$ is unknown). The fact that the global target modeller (Fig.\ \ref{graphical_model}(c)) processes $\mathsf{F}_{\mathsf{S}}$ (a result of learning) is an important progression beyond conventional global multitask learning (Fig.\ \ref{graphical_model}(b)), and we reserve the term \emph{Bayesian transfer learning} for the case in (Fig.\ \ref{graphical_model}(c)). Among its consequences are the following:
\begin{itemize}[leftmargin=16pt]
	\item[(i)] the target learns only from the sufficient statistics of the source, without itself receiving the source data, $\mathbf{y}_{\mathsf{S}}$ (of course, it also processes its local target data, $\mathbf{y}_{\mathsf{T}}$); and
	\item[(ii)] the target can be enriched by the (possibly expert) local source model.
\end{itemize}

It remains to decide on the optimal form of the target's $\mathsf{F}_{\mathsf{S}}$-conditioned joint model,  $\mathsf{M}(\cdot|\mathsf{F}_{\mathsf{S}},\mathbf{y}^{}_{\mathsf{T}})$. Here, $\mathsf{F}_{\mathsf{S}}\equiv\mathsf{F}_{\mathsf{S}}(\mathbf{y}^{\star}_{\mathsf{S}}|\mathbf{y}^{}_{\mathsf{S}})$, i.e.\ the source transfers its source-data-conditioned predictor of an unrealized (i.e.\ unobserved) output, denoted by $\mathbf{y}^{\star}_{\mathsf{S}}$, for which the known input is $\mathbf{x}^{\star}_{\mathsf{S}}$. Hence, the target's unspecified joint model must be augmented, to yield
\begin{equation}
	\mathsf{M}(\mathbf{y}^{\star}_{\mathsf{S}},\mathbf{f}^{}_{\mathsf{S}},\mathbf{f}^{}_{\mathsf{T}}|\mathsf{F}_{\mathsf{S}},\mathbf{y}^{}_{\mathsf{T}})
	=
	\mathsf{M}(\mathbf{y}^{\star}_{\mathsf{S}}|\mathbf{f}^{}_{\mathsf{S}},\mathbf{f}^{}_{\mathsf{T}},\mathsf{F}_{\mathsf{S}},\mathbf{y}^{}_{\mathsf{T}})\mathsf{M}(\mathbf{f}^{}_{\mathsf{S}},\mathbf{f}^{}_{\mathsf{T}}|\mathsf{F}_{\mathsf{S}},\mathbf{y}^{}_{\mathsf{T}})
	,
	\label{unknown_joint_model}
\end{equation}
which follows from the chain rule. Fully probabilistic design (FPD) is the axiomatically justified Bayesian decision-making framework \cite{karny1996towards,karny2012axiomatisation} for optimally choosing the required conditional distribution. FPD is closely related to the minimum cross-entropy principle \cite{shore1980axiomatic}, as further explained in \cite{quinn2016fully}. It solves the following constrained optimization problem:
\begin{equation}
	\mathsf{M}^{\mathsf{o}}(\mathbf{y}^{\star}_{\mathsf{S}},\mathbf{f}^{}_{\mathsf{S}},\mathbf{f}^{}_{\mathsf{T}}|\mathsf{F}_{\mathsf{S}},\mathbf{y}^{}_{\mathsf{T}})
	\equiv
	\underset{\mathsf{M}\in\bm{\mathsf{M}}}{\operatorname{argmin}}\,\mathcal{D}(\mathsf{M}||\mathsf{M}_{\mathsf{I}})
	,
	\label{fpd_optimization_problem}
\end{equation}
where
\begin{equation}
	\mathcal{D}(\mathsf{M}||\mathsf{M}_{\mathsf{I}})
	=
	\mathsf{E}_{\mathsf{M}}\left[\log\left(\frac{\mathsf{M}}{\mathsf{M}_{\mathsf{I}}}\right)\right]
	,
	\nonumber
\end{equation}
is the Kullback-Leibler divergence \cite{kullback1951information} from $\mathsf{M}$ to $\mathsf{M}_{\mathsf{I}}$, and $\mathsf{E}_{\mathsf{M}}$ is the expected value under $\mathsf{M}$. The FPD-optimal model \eqref{fpd_optimization_problem} is the solution (i) that respects the knowledge constraints given by the set of possible models, $\mathsf{M}\in\bm{\mathsf{M}}$, and (ii) that is closest to the ideal model, $\mathsf{M}_{\mathsf{I}}$. The latter must declare the target's preferences about $\mathsf{M}$. These are as follows:

\begin{itemize}[leftmargin=16pt]
	\item[(i)] The set of possible models, $\bm{\mathsf{M}}$, is delineated by restricting the functional form of \eqref{unknown_joint_model},
	\begin{equation}
		\mathsf{M}(\mathbf{y}^{\star}_{\mathsf{S}}|\mathbf{f}^{}_{\mathsf{S}},\mathbf{f}^{}_{\mathsf{T}},\mathsf{F}_{\mathsf{S}},\mathbf{y}^{}_{\mathsf{T}})
		\equiv
		\mathsf{F}(\mathbf{y}_{\mathsf{S}}^{\star}|\mathbf{f}^{}_{\mathsf{S}})
		,
		\label{unknown_model_constraint}
	\end{equation}
	where we assume that $\mathbf{y}^{\star}_{\mathsf{S}}$ is conditionally independent of $(\mathbf{f}^{}_{\mathsf{T}},\mathsf{F}^{}_{\mathsf{S}},\mathbf{y}^{}_{\mathsf{T}})$ given $\mathbf{f}^{}_{\mathsf{S}}$. \eqref{unknown_joint_model} then has the form
	\begin{equation}
		\mathsf{M}(\mathbf{y}^{\star}_{\mathsf{S}},\mathbf{f}^{}_{\mathsf{S}},\mathbf{f}^{}_{\mathsf{T}}|\mathsf{F}_{\mathsf{S}},\mathbf{y}^{}_{\mathsf{T}})
		\equiv
		\mathsf{F}(\mathbf{y}_{\mathsf{S}}^{\star}|\mathbf{f}^{}_{\mathsf{S}})\mathsf{M}(\mathbf{f}^{}_{\mathsf{S}},\mathbf{f}^{}_{\mathsf{T}}|\mathsf{F}_{\mathsf{S}},\mathbf{y}^{}_{\mathsf{T}})
		,
		\label{restricted_unknown_joint_model}
	\end{equation}
	where the $\mathsf{F}_{\mathsf{S}}$-conditioned factor remains unknown and unrestricted. Since $\mathsf{F}(\mathbf{y}_{\mathsf{S}}^{\star}|\mathbf{f}^{}_{\mathsf{S}})$ is fixed, then $\mathsf{M}(\mathbf{f}^{}_{\mathsf{S}},\mathbf{f}^{}_{\mathsf{T}}|\mathsf{F}_{\mathsf{S}},\mathbf{y}^{}_{\mathsf{T}})$ is the only unknown quantity in \eqref{restricted_unknown_joint_model}, and the set of possible models is therefore
	\begin{align}
		\mathsf{M}\in\bm{\mathsf{M}}
		&\equiv
		\lbrace\text{models \eqref{restricted_unknown_joint_model} with $\mathsf{F}(\mathbf{y}_{\mathsf{S}}^{\star}|\mathbf{f}^{}_{\mathsf{S}})$ fixed}
		\nonumber
		\\
		&\hspace{35pt}
		\text{and $\mathsf{M}(\mathbf{f}^{}_{\mathsf{S}},\mathbf{f}^{}_{\mathsf{T}}|\mathsf{F}_{\mathsf{S}},\mathbf{y}^{}_{\mathsf{T}})$ variational}\rbrace
		.
		\label{knowledge_constrained_set}
	\end{align}
	\item[(ii)] Adopting the same chain rule expansion as in \eqref{unknown_joint_model}, the target's ideal model is specified as
	\begin{equation}
		\mathsf{M}_{\mathsf{I}}(\mathbf{y}^{\star}_{\mathsf{S}},\mathbf{f}^{}_{\mathsf{S}},\mathbf{f}^{}_{\mathsf{T}}|\mathsf{F}_{\mathsf{S}},\mathbf{y}^{}_{\mathsf{T}})
		\equiv
		\mathsf{F}_{\mathsf{S}}(\mathbf{y}_{\mathsf{S}}^{\star}|\mathbf{y}^{}_{\mathsf{S}})\mathsf{F}(\mathbf{f}^{}_{\mathsf{S}},\mathbf{f}^{}_{\mathsf{T}}|\mathbf{y}^{}_{\mathsf{T}})
		.
		\label{ideal_model}
	\end{equation}
	Comparing with \eqref{unknown_joint_model}, the following ideal declarations have been made by the target:
	\begin{align}
		\mathsf{M}_{\mathsf{I}}(\mathbf{y}^{\star}_{\mathsf{S}}|\mathbf{f}^{}_{\mathsf{S}},\mathbf{f}^{}_{\mathsf{T}},\mathsf{F}_{\mathsf{S}},\mathbf{y}^{}_{\mathsf{T}})
		&\equiv
		\mathsf{F}_{\mathsf{S}}(\mathbf{y}_{\mathsf{S}}^{\star}|\mathbf{y}^{}_{\mathsf{S}})
		,
		\label{ideal_model_constraint_a}
		\\
		\mathsf{M}_{\mathsf{I}}(\mathbf{f}^{}_{\mathsf{S}},\mathbf{f}^{}_{\mathsf{T}}|\mathsf{F}_{\mathsf{S}},\mathbf{y}^{}_{\mathsf{T}})
		&\equiv
		\mathsf{F}(\mathbf{f}^{}_{\mathsf{S}},\mathbf{f}^{}_{\mathsf{T}}|\mathbf{y}^{}_{\mathsf{T}})
		.
		\label{ideal_model_constraint_b}
	\end{align}
	Specifically, in \eqref{ideal_model_constraint_a}, the target assumes as its ideal that $\mathbf{y}^{\star}_{\mathsf{S}}$ is conditionally independent of $(\mathbf{f}^{}_{\mathsf{S}},\mathbf{f}^{}_{\mathsf{T}},\mathbf{y}^{}_{\mathsf{T}})$ given $\mathsf{F}^{}_{\mathsf{S}}$, and---in order to transfer the source output-data predictor, $\mathsf{F}_{\mathsf{S}}$---adopts the source's predictor of $\mathbf{y}^{\star}_{\mathsf{S}}$, i.e.\ $\mathsf{F}_{\mathsf{S}}(\mathbf{y}_{\mathsf{S}}^{\star}|\mathbf{y}^{}_{\mathsf{S}})$, as its ideal model for $\mathbf{y}^{\star}_{\mathsf{S}}$. Similarly, in \eqref{ideal_model_constraint_b}, the target ideally models $(\mathbf{f}^{}_{\mathsf{S}},\mathbf{f}^{}_{\mathsf{T}})$ conditionally independently of $\mathsf{F}_{\mathsf{S}}$ given its own $\mathbf{y}^{}_{\mathsf{T}}$, and adopts its own joint model, $\mathsf{F}(\mathbf{f}^{}_{\mathsf{S}},\mathbf{f}^{}_{\mathsf{T}}|\cdot)$, as its ideal, $\mathsf{M}_{\mathsf{I}}(\mathbf{f}^{}_{\mathsf{S}},\mathbf{f}^{}_{\mathsf{T}}|\cdot)$.
\end{itemize}
These specifications imply a unique design for the target's source-knowledge-constrained optimal distribution \eqref{fpd_optimization_problem}, as established in the following proposition.

\begin{prop}\label{fpd_optimal_static_transfer}
	The target modeller constrains the unknown model \eqref{unknown_joint_model} to belong to the set of possible models, $\mathsf{M}\in\bm{\mathsf{M}}$ \eqref{knowledge_constrained_set}, and adopts the ideal model $\mathsf{M}_{\mathsf{I}}$ \eqref{ideal_model}. Then the target's FPD-optimal model \eqref{fpd_optimization_problem} is
	\begin{align}
		\mathsf{M}^{\mathsf{o}}(\mathbf{y}^{\star}_{\mathsf{S}},\mathbf{f}^{}_{\mathsf{S}},\mathbf{f}^{}_{\mathsf{T}}|\mathsf{F}_{\mathsf{S}},\mathbf{y}^{}_{\mathsf{T}})
		&=
		\mathsf{F}(\mathbf{y}_{\mathsf{S}}^{\star}|\mathbf{f}^{}_{\mathsf{S}})\mathsf{M}^{\mathsf{o}}(\mathbf{f}^{}_{\mathsf{S}},\mathbf{f}^{}_{\mathsf{T}}|\mathsf{F}_{\mathsf{S}},\mathbf{y}^{}_{\mathsf{T}})
		,
		\label{fpd_optimal_model}
	\end{align}
	where
	\begin{align}
		\mathsf{M}^{\mathsf{o}}(\mathbf{f}^{}_{\mathsf{S}},\mathbf{f}^{}_{\mathsf{T}}|\mathsf{F}_{\mathsf{S}},\mathbf{y}^{}_{\mathsf{T}})
		&\propto
		\mathsf{F}(\mathbf{f}^{}_{\mathsf{S}},\mathbf{f}^{}_{\mathsf{T}}|\mathbf{y}^{}_{\mathsf{T}})
		\nonumber
		\\
		&\hspace{-24pt}\times
		\exp\big\lbrace-\mathcal{D}\big(\mathsf{F}(\mathbf{y}^{\star}_{\mathsf{S}}|\mathbf{f}^{}_{\mathsf{S}})||\mathsf{F}_{\mathsf{S}}(\mathbf{y}_{\mathsf{S}}^{\star}|\mathbf{y}^{}_{\mathsf{S}})\big)\big\rbrace
		.
		\label{fpd_optimal_posterior_distribution_generic}
	\end{align}
	\begin{proof}
		See Appendix \ref{fpd_optimal_static_transfer_proof}.
	\end{proof}
\end{prop}

The design \eqref{fpd_optimal_posterior_distribution_generic} is the optimal Bayesian conditioning mechanism we have been looking for. It forms the update from the pre-posterior, $\mathsf{F}(\mathbf{f}^{}_{\mathsf{S}},\mathbf{f}^{}_{\mathsf{T}}|\mathbf{y}^{}_{\mathsf{T}})$, to the FPD-optimal, $\mathsf{F}_{\mathsf{S}}$-conditioned, posterior, $\mathsf{M}^{\mathsf{o}}(\mathbf{f}^{}_{\mathsf{S}},\mathbf{f}^{}_{\mathsf{T}}|\mathsf{F}_{\mathsf{S}},\mathbf{y}^{}_{\mathsf{T}})$. The exponential structure in \eqref{fpd_optimal_posterior_distribution_generic} constitutes the updating term which incorporates $\mathsf{F}_{\mathsf{S}}$. It has the Boltzmann structure that is typical of entropic designs \cite{quinn2016fully,quinn2017optimal}.

\SetAlgoVlined
\SetAlFnt{\footnotesize}
\SetAlCapNameFnt{\footnotesize}
\SetAlCapFnt{\footnotesize}
\RestyleAlgo{ruled}
\SetAlCapHSkip{0pt}
\SetAlgoSkip{-1pt}
\SetInd{2pt}{3pt}
\SetNlSkip{3pt}
\DecMargin{9pt}
\SetAlgoHangIndent{0pt}

\SetKwInput{KwSet}{set}

\begin{algorithm}[t]
	\setstretch{1.1}
	\caption{The FPD-optimal global target GP regression modeller}
	\label{FPD}

	\KwIn{$k_{\mathsf{SS}}$, $k_{\mathsf{ST}}$, $k_{\mathsf{TT}}$, $\mathbf{x}_{\mathsf{S}}$, $\mathbf{x}_{\mathsf{T}}$, $\bar{\mathbf{m}}_{\mathsf{S}}$, $\bar{\mathbf{R}}_{\mathsf{S}}$, $\mathbf{y}_{\mathsf{T}}$, $\sigma^{2}_{\mathsf{T}}$}

	For all entries of $\mathbf{x}$, compute
	\vspace{-5pt}
	\begin{mynumbering}
		\begin{subequations}\label{optimal_posterior_gp_functions}
			\begin{align}
				\hspace{-10pt}m_{q}^{\mathsf{o}}(x)
				&=
				\mathbf{k}_{q}(\mathbf{K}+\operatorname{blkdiag}(\bar{\mathbf{R}}_{\mathsf{S}},\sigma^{2}_{\mathsf{T}}I_{n}))^{-1}
				\left[
					\begin{matrix}
						\bar{\mathbf{m}}_{\mathsf{S}}
						\\
						\mathbf{y}_{\mathsf{T}}
					\end{matrix}
				\right]
				,
				\label{optimal_posterior_gp_mean_function}
				\\[4pt]
				\hspace{-10pt}k_{qp}^{\mathsf{o}}(x,x')
				&=
				k_{qp}(x,x')-\mathbf{k}_{q}(\mathbf{K}+\operatorname{blkdiag}(\bar{\mathbf{R}}_{\mathsf{S}},\sigma^{2}_{\mathsf{T}}I_{n}))^{-1}\mathbf{k}'_{p}
				,
				\label{optimal_posterior_gp_covariance_function}
			\end{align}
		\end{subequations}
		where
		\begin{subequations}\label{interaction_structure}
			\begin{align}
				&\mathbf{k}_{q}
				\equiv
				\left[
					\setlength\arraycolsep{2pt}
					\begin{matrix}
						k_{q\mathsf{S}}(x,\mathbf{x}_{\mathsf{S}}) & k_{q\mathsf{T}}(x,\mathbf{x}_{\mathsf{T}})
					\end{matrix}
				\right]
				,
				\hspace{15pt}
				\mathbf{k}'_{p}
				\equiv
				\left[
					\begin{matrix}
						k_{\mathsf{S}p}(\mathbf{x}_{\mathsf{S}},x') \\ k_{\mathsf{T}p}(\mathbf{x}_{\mathsf{T}},x')
					\end{matrix}
				\right]
				,
				\\
				&\hspace{40pt}\mathbf{K}
				\equiv
				\left[
					\setlength\arraycolsep{2pt}
					\begin{matrix}
						k_{\mathsf{SS}}(\mathbf{x}_{\mathsf{S}},\mathbf{x}_{\mathsf{S}}) & k_{\mathsf{ST}}(\mathbf{x}_{\mathsf{S}},\mathbf{x}_{\mathsf{T}})\\
						k_{\mathsf{TS}}(\mathbf{x}_{\mathsf{T}},\mathbf{x}_{\mathsf{T}}) & k_{\mathsf{TT}}(\mathbf{x}_{\mathsf{T}},\mathbf{x}_{\mathsf{T}})
					\end{matrix}
				\right]
				.
			\end{align}
		\end{subequations}
	\end{mynumbering}
	\hspace{-3pt}Here, $q,p\in(\mathsf{S},\mathsf{T})$, and $\operatorname{blkdiag}(U,V)$ is the block diagonal matrix obtained from $U\in\mathbb{R}^{u\times u}$ and $V\in\mathbb{R}^{v\times v}$.
	
	\KwOut{$\mathbf{m}_{q}^{\mathsf{o}}\equiv m_{q}^{\mathsf{o}}(\mathbf{x}_{q})$, $\mathbf{K}_{qp}^{\mathsf{o}}\equiv k_{qp}^{\mathsf{o}}(\mathbf{x}_{q},\mathbf{x}_{p})$}
\end{algorithm}

\section{Local and global Gaussian process\\regression modellers}\label{local_and_global_gpr_modellers}
We instantiate the framework in Section \ref{btl_for_global_modellers} to the case of source and target GP regression tasks \cite{bishop2006pattern,murphy2012machine}. For this purpose, consider a stochastic process, $f\sim\mathsf{F}$, over a scalar random function, $f\in\mathbb{R}$, such that any $n$-dimensional collection of evaluation points (i.e.\ regressors), $\mathbf{x}$, induces a joint probability distribution over the function values, $\mathbf{f}$. The GP, $f(x)\sim\mathcal{GP}(m_{\theta_{f}}(x),k_{\theta_{f}}(x,x'))$, is conditioned on known mean function, $m_{\theta_{f}}:\mathbb{R}^{n_{x}}\rightarrow\mathbb{R}$, assumed (without loss of generality) to be zero. The known covariance (kernel) function is $k_{\theta_{f}}:\mathbb{R}^{n_{x}\times n_{x}}\rightarrow\mathbb{R}$. In this way, all joint distributions are specified to be multivariate Gaussian, $\mathsf{F}(\mathbf{f})\equiv\mathcal{N}(\mathbf{m},\mathbf{K})$. Here, $\mathbf{m}\equiv m_{\theta_{f}}(\mathbf{x})$ and $\mathbf{K}\equiv k_{\theta_{f}}(\mathbf{x},\mathbf{x})$ are the known, $n$-dimensional mean vector, and $n\times n$-dimensional covariance matrix, respectively. We suppress $\theta_{f}$---the parameters of $m_{\theta_{f}}$ and $k_{\theta_{f}}$---in the sequel for notational simplicity.

This GP, $f$, is indirectly (i.e.\ noisily) observed as $y$ at the evaluation points, $\mathbf{x}$, via an additive, uncorrelated, Gaussian noise process. Hence, the joint distribution is
\begin{subequations}\label{single_task_regression}
	\begin{align}
		\mathsf{F}(\mathbf{y}|\mathbf{f},\theta)
		&=
		\mathcal{N}(\mathbf{f},\sigma^{2}I_{n})
		,
		\\
		\mathsf{F}(\mathbf{f}|\theta)
		&=
		\mathcal{N}(\mathbf{m},\mathbf{K})
		,
		\label{gp_prior}
	\end{align}
\end{subequations}
with the GP as prior \eqref{gp_prior}. Here, $\sigma^{2}$ is the conditional output-data variance, $I_{n}$ is the $n\times n$-dimensional identity matrix, and $\theta\equiv(\theta_{f},\sigma^{2})$ are all known parameters. In the rest of this section, we restore the explicit conditioning on $\theta$, as it is relevant to the calculus.

We now adopt this stochastic structure in the isolated (i.e.\ unconditioned on transferred knowledge) local source task of BTL (Fig.\ \ref{graphical_model}(c)). Our purpose is to transfer an output-data predictor via construction of the GP posterior at unobserved test points, as follows.
\begin{rem}\label{source_gp_regression_remark}
	The isolated local source modeller adopts $\mathsf{F}_{\mathsf{S}}(\mathbf{y}_{\mathsf{S}},\mathbf{f}_{\mathsf{S}}|\theta_{\mathsf{S}})$ \eqref{single_task_regression}, which is assumed to be strict-sense stationary. Then, the posterior distribution---evaluated at a single test point, $x\in\mathbb{R}^{n_{x}}$ (now shown explicitly in the conditioning)---is
	\begin{equation}
		\mathsf{F}_{\mathsf{S}}(f_{\mathsf{S}}|\mathbf{y}_{\mathsf{S}},\theta_{\mathsf{S}},x)
		=
		\mathcal{N}(\bar{m}_{\mathsf{S}},\bar{k}_{\mathsf{S}})
		,
		\nonumber
	\end{equation}
	where
	\begin{subequations}\label{posterior_gp_functions}
		\begin{align}
			\bar{m}_{\mathsf{S}}(x)
			&=
			\mathbf{k}_{\mathsf{S}}(\mathbf{K}_{\mathsf{S}}+\sigma_{\mathsf{S}}^{2}I_{n})^{-1}\mathbf{y}_{\mathsf{S}}
			,
			\label{source_posterior_mean_function}
			\\
			\bar{k}_{\mathsf{S}}(x,x')
			&=
			k_{\mathsf{S}}(x,x')-\mathbf{k}_{\mathsf{S}}(\mathbf{K}_{\mathsf{S}}+\sigma_{\mathsf{S}}^{2}I_{n})^{-1}\mathbf{k}'_{\mathsf{S}}
			,
		\end{align}
	\end{subequations}
	with $\mathbf{k}_{\mathsf{S}}\equiv k_{\mathsf{S}}(x,\mathbf{x}_{\mathsf{S}})$, $\mathbf{k}'_{\mathsf{S}}\equiv k_{\mathsf{S}}(\mathbf{x}_{\mathsf{S}},x')$, and $\mathbf{K}_{\mathsf{S}}\equiv k_{\mathsf{S}}(\mathbf{x}_{\mathsf{S}},\mathbf{x}_{\mathsf{S}})$. Here, $(x,x')$ is a second order test pair. The posterior predictor of unobserved outputs, $\mathbf{y}_{\mathsf{S}}^{\star}$, for which the (known) test points are $\mathbf{x}_{\mathsf{S}}^{\star}$, is
	\begin{equation}
		\mathsf{F}_{\mathsf{S}}(\mathbf{y}_{\mathsf{S}}^{\star}|\mathbf{y}_{\mathsf{S}},\theta_{\mathsf{S}})
		=
		\mathcal{N}(\bar{\mathbf{m}}_{\mathsf{S}},\bar{\mathbf{R}}_{\mathsf{S}})
		,
		\label{posterior_predictor}
	\end{equation}
	where $\bar{\mathbf{R}}_{\mathsf{S}}=\bar{\mathbf{K}}_{\mathsf{S}}+\sigma^{2}_{\mathsf{S}}I_{n^{\star}}$.
	\begin{proof}
		See, for example, \cite{rasmussen2006gaussian}, Chapter 2.
	\end{proof}
\end{rem}

Recall that only the source is a local modeller of the kind above. In contrast, the target modeller augments its model to assess the global source-target system (Fig.\ \ref{graphical_model}(c)). Therefore, the joint prior distribution, $\mathsf{F}(\mathbf{y}^{\star}_{\mathsf{S}},\mathbf{y}^{}_{\mathsf{T}},\mathbf{f}^{}_{\mathsf{S}},\mathbf{f}^{}_{\mathsf{T}}|\theta)$, adopted by the global target modeller before processing $\mathbf{y}^{}_{\mathsf{T}}$ and $\mathsf{F}_{\mathsf{S}}$, is specified~by
\begin{subequations}\label{ideal_model_specific}
	\begin{align}
		\mathsf{F}(\mathbf{y}^{\star}_{\mathsf{S}}|\mathbf{f}^{}_{\mathsf{S}},\theta)
		&\equiv
		\mathcal{N}(\mathbf{f}^{}_{\mathsf{S}},(\sigma^{\star})^{2}I_{n^{\star}})
		,
		\label{source_output_data_model}
		\\
		\mathsf{F}(\mathbf{y}^{}_{\mathsf{T}}|\mathbf{f}^{}_{\mathsf{T}},\theta)
		&\equiv
		\mathcal{N}(\mathbf{f}^{}_{\mathsf{T}},\sigma_{\mathsf{T}}^{2}I_{n})
		,
		\label{target_output_data_model}
		\\
		\mathsf{F}(\mathbf{f}^{}_{\mathsf{S}},\mathbf{f}^{}_{\mathsf{T}}|\theta)
		&\equiv
		\mathcal{N}(\mathbf{0},\mathbf{K})
		,
		\label{pre_prior}
	\end{align}
\end{subequations}
where
\begin{equation}
	\mathbf{K}
	=
	\left[
		\setlength\arraycolsep{2pt}
		\begin{matrix}
			\mathbf{K}^{}_{\mathsf{S}\mathsf{S}} & \mathbf{K}^{}_{\mathsf{S}\mathsf{T}}
			\\
			\mathbf{K}^{T}_{\mathsf{S}\mathsf{T}} & \mathbf{K}^{}_{\mathsf{T}\mathsf{T}}
		\end{matrix}
	\right]
	\label{pre_prior_covariance_matrix}
\end{equation}
is the symmetric covariance matrix describing the target's prior beliefs about the interactions between the source and target GP function values, $(\mathbf{f}^{}_{\mathsf{S}},\mathbf{f}^{}_{\mathsf{T}})$. It is expressed in a block-matrix form, involving three matrix degrees-of-freedom ($T$ denotes matrix transposition). Specific cases will be considered in Section \ref{synthetic_data_experiments}. Also, $(\sigma^{\star})^{2}$ is the target's conditional variance of the source output data at any test point, and $\sigma_{\mathsf{T}}^{2}$ is the target's conditional variance of any target output datum.

\begin{figure*}[ht!]
	\includegraphics{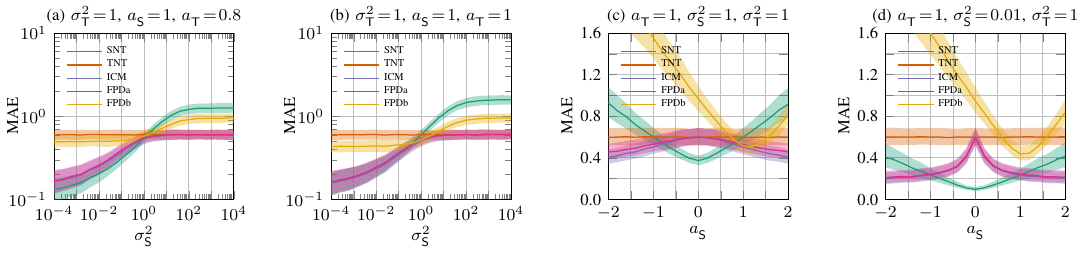}
	\vspace{-20pt}
	\caption{
		\textit{Synthetic data experiments}. The MAE \eqref{mae} vs.\ the conditional output-data variance of the source, $\sigma_{\mathsf{S}}^{2}$, for the source weight (a) $a_{\mathsf{T}}=0.8$, and (b) $a_{\mathsf{T}}=1$. The MAE vs.\ the source weight, $a_{\mathsf{S}}$, for the conditional output-data variance of the source (c) $\sigma_{\mathsf{S}}^{2}=1$ and (d) $\sigma_{\mathsf{S}}^{2}=0.01$. The results are averaged over $5000$ independent simulation runs, with the solid line and the shaded area being the median and the interquartile range, respectively.
	}
	\label{mae_vs_a}
\end{figure*}

It remains to compute the target's posterior joint inference of the GP function values,  $(\mathbf{f}^{}_{\mathsf{S}},\mathbf{f}^{}_{\mathsf{T}})$, under its joint model \eqref{ideal_model_specific}, i.e.\ after it has processed its local data, $\mathbf{y}^{}_{\mathsf{T}}$, and the transferred knowledge, $\mathsf{F}_{\mathsf{S}}$, respectively.
\begin{prop}\label{joint_posterior_model_specific}
	The global target modeller elicits the joint prior distribution \eqref{ideal_model_specific} and processes the source output-data predictor \eqref{posterior_predictor}. Then, the FPD-optimal posterior distribution \eqref{fpd_optimal_posterior_distribution_generic}---evaluated at a test point $x$---is
	\begin{equation}
		\mathsf{M}^{\mathsf{o}}(f^{}_{\mathsf{S}},f^{}_{\mathsf{T}}|\mathsf{F}_{\mathsf{S}},\mathbf{y}^{}_{\mathsf{T}},\theta,x)
		=
		\mathcal{N}(m^{\mathsf{o}},k^{\mathsf{o}})
		,
		\label{fpd_optimal_posterior_distribution_specific}
	\end{equation}
	where
	\begin{equation}
		m^{\mathsf{o}}
		=
		\left[
			\begin{matrix}
				m^{\mathsf{o}}_{\mathsf{S}}(x)
				\\
				m^{\mathsf{o}}_{\mathsf{T}}(x)
			\end{matrix}
		\right]
		,
		\hspace{15pt}
		k^{\mathsf{o}}
		=
		\left[
			\setlength\arraycolsep{2pt}
			\begin{matrix}
				k^{\mathsf{o}}_{\mathsf{S}\mathsf{S}}(x,x') & k^{\mathsf{o}}_{\mathsf{S}\mathsf{T}}(x,x')
				\\
				k^{\mathsf{o}}_{\mathsf{T}\mathsf{S}}(x,x') & k^{\mathsf{o}}_{\mathsf{T}\mathsf{T}}(x,x')
			\end{matrix}
		\right]
		,
		\nonumber
	\end{equation}
	with $m_{q}^{\mathsf{o}}$ and $k_{qp}^{\mathsf{o}}$ being specified in \eqref{optimal_posterior_gp_functions} for $q,p\in(\mathsf{S},\mathsf{T})$.
	\begin{proof}
		See Appendix \ref{joint_posterior_model_specific_proof}.
	\end{proof}
\end{prop}

The proposed global target GP regression modeller, supported by FPD-optimal BTL from the local source GP regression modeller, is summarized in Algorithm~\ref{FPD}.

\section{Synthetic Data Experiments}\label{synthetic_data_experiments}
In this section, we investigate the performance of our algorithm (Algorithm~\ref{FPD}) against a number of alternative transfer learning algorithms, in particular focussing on distinctions between the \emph{analysis model} underlying each algorithm and the \emph{synthesis model} used to generate the synthetic data. We illustrate the following key properties of Algorithm~\ref{FPD}:
\begin{itemize}[leftmargin=12pt]
	\item the ability to achieve robust knowledge transfer---i.e.\ rejection of imprecise source knowledge---due to successful transfer of all moments of $\mathsf{F}_{\mathsf{S}}$ \eqref{posterior_predictor};
	\item the ability to process known correlation between the source and target latent functions, $(\mathbf{f}^{}_{\mathsf{S}},\mathbf{f}^{}_{\mathsf{T}})$, via the target's specification of the covariance structure, $\mathbf{K}$ \eqref{pre_prior_covariance_matrix};
	\item the experimental bounding of the performance of Algorithm~\ref{FPD}, which holds under the ideal condition when the synthesis and analysis models are equal; and
	\item transfer of source model structure via $\mathsf{F}_{\mathsf{S}}$ \eqref{posterior_predictor}, leading to improved positive transfer of Algorithm~\ref{FPD} under mismatch between the source and target models.
\end{itemize}
The first three issues are explored in Section \ref{performance_bound_section}, and the fourth in Section \ref{source_model_structure_section}.
\begin{table}[t]
	\centering
	\footnotesize
	\begin{tabular}{ll|ll}
		\hline
		Algorithm & & Description \\
		\hline
		{\em S}ource {\em N}o {\em T}ransfer & ({\em SNT}) & Remark \ref{source_gp_regression_remark} & \hspace{-5pt}(Fig.\ \ref{graphical_model}a)\\
		{\em T}arget {\em N}o {\em T}ransfer & ({\em TNT}) & Remark \ref{source_gp_regression_remark} & \hspace{-5pt}(Fig.\ \ref{graphical_model}a)\\
		{\em I}ntrinsic {\em C}oregionalization {\em M}odel & ({\em ICM}) & \cite{alvarez2011computationally} & \hspace{-5pt}(Fig.\ \ref{graphical_model}b)\\
		{\bf{\em F}ully {\em P}robabilistic {\em D}esign} & {\bf({\em FPDa})} & {\bf Algorithm~\ref{FPD}} & \hspace{-5pt}{\bf (Fig.\ \ref{graphical_model}c)}\\
		{\em F}ully {\em P}robabilistic {\em D}esign & ({\em FPDb}) & \cite{papez2019bayesian} & \\
		\hline
	\end{tabular}
	\caption{Algorithms compared in Sections \ref{synthetic_data_experiments} and \ref{real_data_experiments}. FPDa is the newly proposed BTL algorithm.}
	\label{algorithms}
\end{table}
Throughout, the synthesis model is the (rank-1) intrinsic coregionalization model (ICM) \cite{alvarez2011computationally},
\begin{align}
	u
	&\sim
	\mathcal{GP}(0,k_{\theta_{u}})
	,
	\nonumber
	\\
	y_{\mathsf{S}}
	&\sim
	\mathcal{N}(a_{\mathsf{S}}u(x),\sigma^{2}_{\mathsf{S}})
	,
	\label{synthesis_model}
	\\
	y_{\mathsf{T}}
	&\sim
	\mathcal{N}(a_{\mathsf{T}}u(x),\sigma^{2}_{\mathsf{T}})
	,
	\nonumber
\end{align}
and so $x_{\mathsf{S}}=x_{\mathsf{T}}=x$. Here, $f_{\mathsf{S}}(x)=a_{\mathsf{S}}u(x)$ and $f_{\mathsf{T}}(x)=a_{\mathsf{T}}u(x)$. It follows that the covariance matrix \eqref{pre_prior_covariance_matrix} is
\begin{equation}
	\mathbf{K}
	=
	B\otimes k_{\theta_{u}}(\mathbf{x},\mathbf{x})
	,
	\label{pre_prior_covariance_matrix_specific}
\end{equation}
where
\begin{equation}
	B
	=
	\left[
		\begin{matrix}
			a_{\mathsf{S}}
			\\
			a_{\mathsf{T}}
		\end{matrix}
	\right]
	\left[
		\begin{matrix}
			a_{\mathsf{S}}
			\\
			a_{\mathsf{T}}
		\end{matrix}
	\right]^{T}
	\equiv
	\left[
		\setlength\arraycolsep{2pt}
		\begin{matrix}
			b_{\mathsf{SS}} & b_{\mathsf{ST}}
			\\
			b_{\mathsf{ST}} & b_{\mathsf{TT}}
		\end{matrix}
	\right]
	\in
	\mathbb{R}^{2\times 2}
	\label{coregionalization_matrix}
\end{equation}
is the coregionalization matrix \cite{alvarez2012kernels}, and $\otimes$ denotes the Kronecker product of the matrix arguments. We adopt the mean absolute error (MAE) as the performance measure:
\begin{equation}\label{mae}
	\text{MAE}=\frac{1}{n^{\ast}}\sum^{n^{\ast}}_{i=1}|f_{\mathsf{T}}(x^{\ast}_{i})-m_{\mathsf{T}}^{\mathsf{o}}(x^{\ast}_{i})|
	,
\end{equation}
where $f_{\mathsf{T}}(x_{i}^{\ast})$ is the true target function value (Fig.\ \ref{graphical_model}), $m^{\mathsf{o}}_{\mathsf{T}}(x_{i}^{\ast})$ is the lower sub-vector of the posterior predictive mean estimate \eqref{optimal_posterior_gp_mean_function}, $|\cdot|$ is the absolute value, and $(x^{\ast}_{i})^{n^{\ast}}_{i=1}$ are the $n^{\ast}\geq 1$ test points.

\begin{figure*}[t]
	\includegraphics{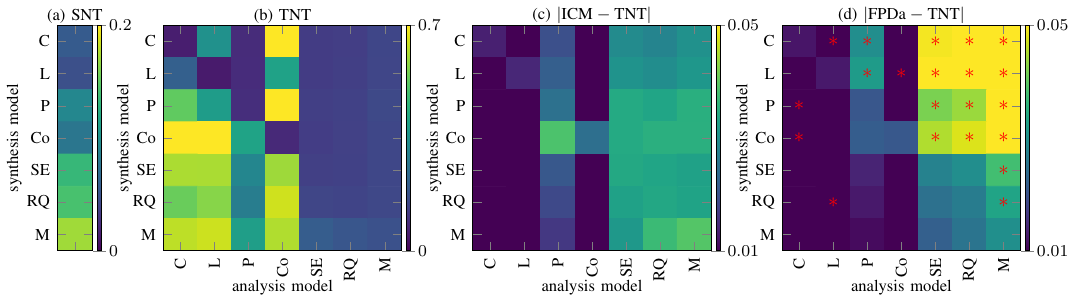}
	\vspace{-15pt}
	\caption{
		\textit{Synthetic data experiments}. (a) The MAE \eqref{mae} of the SNT algorithm utilizing the analysis model which is equivalent to the source part of the ICM synthesis model. (b) The MAE of the TNT algorithm for all combinations of its analysis model kernel and the ICM synthesis model kernel (Table \ref{kernels}). In (a) and (b), \emph{lower} is better. (c) and (d): The \emph{absolute} value of differential MAE of the ICM algorithm and the FPDa algorithm, respectively, for all combinations of analysis model kernel and ICM synthesis model kernel (the absolute differential is with respect to the MAE of the TNT algorithm in (b), for ease of comparison of the positive transfer attained  (always) by the respective algorithms). In (c) and (d), \emph{higher} is better. The results are averaged over $1000$ independent simulation runs. The asterisk \textcolor{red}{$*$} highlights the cases where the FPDa algorithm achieves greater positive transfer than the ICM algorithm.
	}
	\label{structure}
\end{figure*}

\begin{table*}[t]
	\centering
	\footnotesize
	\begin{tabular}{ll|l}
		\hline
		Covariance function && Expression \\
		\hline
		{\em C}onstant & (C) & $k(x,x')=\sigma^{2}$ \\
		{\em L}inear & (L)& $k(x,x')=\sigma^{2}xx'$ \\
		{\em P}olynomial & (P) & $k(x,x')=(\sigma^{2}xx'+\gamma)^{d}$ \\
		{\em Co}sine & (Co) & $k(x,x')=\sigma^{2}\cos\big(2\pi\sum_{i=1}^{n_{x}}(x_{i}-x'_{i})/l^{2}\big)$ \\
		{\em S}quared {\em E}xponential & (SE) & $k(x,x')=\sigma^{2}\exp\big\lbrace -\tfrac{1}{2}\tfrac{r(x,x')^{2}}{l^{2}}\big\rbrace$ \\
		{\em R}ational {\em Q}uadratic & (RQ) & $k(x,x')=\sigma^{2}\big(1+\tfrac{1}{2}\tfrac{r(x,x')^{2}}{l^{2}\alpha}\big)^{-\alpha}$ \\
		{\em M}at\'{e}rn ($\nu=3/2$) & (M) & $k(x,x')=\sigma^{2}(1+\sqrt 3r(x,x'))\exp\lbrace -\sqrt 3r(x,x')\rbrace$ \\
		\hline
	\end{tabular}
	\caption{Kernel functions adopted in this paper, see \cite{rasmussen2006gaussian}. Here, $\sigma^{2}$ is the signal variance, $\gamma$ is the offset, $d$ is the polynomial degree, $l^{2}$ is the length-scale, $\alpha$ is the fluctuation parameter, and $r(x,x')\equiv||x-x'||$ is the Euclidean distance between $x$ and $x'$. Note that the Mat\'{e}rn kernel is assessed only in the case of the smoothness parameter $\nu=3/2$.}
	\label{kernels}
\end{table*}

\subsection{The bounding performance of our BTL algorithm (Fig.\ \ref{mae_vs_a})}\label{performance_bound_section}
\textit{Settings}:\ We choose $k_{\theta_{u}}$ in \eqref{pre_prior_covariance_matrix_specific} as the squared exponential (covariance) kernel function with the parameters $\theta_{u}\equiv(\sigma^{2}_{u}, l_{u}^{2})$ (the full list of kernel functions involved in Section \ref{synthetic_data_experiments} are defined in Table \ref{kernels}). The complete parameterization of \eqref{synthesis_model} is therefore $\theta=(\sigma^{2}_{\mathsf{S}},\sigma^{2}_{\mathsf{T}},a_{\mathsf{S}},a_{\mathsf{T}},\sigma^{2}_{u},l_{u}^{2})=(1,1,0.8,1,2,0.4)$. Furthermore, $n=64$ scalar input data (i.e.\ $n_{x}=1$) are generated via $x\sim\mathcal{U}(-3.5,$ $3.5)$, where $\mathcal{U}(a,b)$ is the uniform distribution on the open interval $(a,b)$. The $n^{\ast}=200$ test points, $(x^{\ast}_{i})^{n^{\ast}}_{i=1}$, are placed on a uniform grid in the interval $(-5,5)$.

\textit{Performance}:\ We compare the performance of the algorithms in Table \ref{algorithms} as we adapt the source knowledge quality via $\sigma^{2}_{\mathsf{S}}$ \eqref{synthesis_model} and the correlation between the source and target latent functions via $a_{\mathsf{S}}$ in \eqref{coregionalization_matrix}, holding all other parameters in $\theta$ constant. In this experiment, we eliminate misspecification between the ICM synthesis model \eqref{synthesis_model} and the analysis models of the various algorithms in Table \ref{algorithms}; i.e.\ they all have perfect knowledge of $\theta$. Fig.\ \ref{mae_vs_a}(a) and (b) show the MAE for target weights, $a_{\mathsf{T}}^{}=0.8$ and $1.0$, respectively, as a function of source variance, $\sigma_{\mathsf{S}}^{2}$. The TNT algorithm (i.e.\ isolated target task) defines the baseline performance since it does not depend on the source knowledge, and so it is invariant to all source settings. If any of the BTL algorithms (FPDa and FPDb) yields an MAE below or above this level, it is said to deliver \textit{positive} or \textit{negative} transfer, respectively. If any algorithm saturates at this level, it is said to achieve \textit{robust} transfer. The source MAE of the SNT algorithm (also an isolated task) is computed in terms of the quantities $f_{\mathsf{S}}$ and $m_{\mathsf{S}}^{\mathsf{o}}$, and, therefore, is unrelated to the target MAE \eqref{mae}. Indeed, we present this source MAE to track the influence of changing $\sigma^{2}_{\mathsf{S}}$ and $a_{\mathsf{S}}$ on the performance of the SNT algorithm. The latter provides the transferred source output-data predictor to the FPDa and FPDb algorithms. Consequently, the performance of SNT will impact FPDa and FPDb, as explained in the next paragraph.

The ICM algorithm delimits the optimal performance for all $\sigma^{2}_{\mathsf{S}}$, since it adopts the ICM synthesis model \eqref{synthesis_model} as its analysis model (i.e.\ misspecification is completely eliminated). Our FPDa algorithm (Algorithm \ref{FPD}) very closely follows the performance of the ICM algorithm (Fig.\ \ref{mae_vs_a}), despite (i) adopting incomplete modelling, and (ii) processing source statistics rather than the raw source output data, $y_{\mathsf{S}}$; i.e.\ FPDa is a robust BTL algorithm. This improves on our previous FPDb algorithm, which achieves positive transfer only for $\sigma^{2}_{\mathsf{S}}<\sigma^{2}_{\mathsf{T}}$ (high-quality source knowledge) but negative transfer for $\sigma^{2}_{\mathsf{S}}>\sigma^{2}_{\mathsf{T}}$ (low-quality source knowledge); i.e.\ it is non-robust.

Fig.\ \ref{mae_vs_a} depicts the MAE for fixed target weight $a_{\mathsf{T}}=1$, and varying source weight, $a_{\mathsf{S}}$ (Fig.\ \ref{mae_vs_a}(c) and (d)). The FPDa algorithm again achieves close tracking of the ICM algorithm for all the parameter settings. We note that the critical point where the performance curves converge occurs when $\sigma_{\mathsf{S}}^{2}=\sigma_{\mathsf{T}}^{2}=1$. For this setting, we explore the influence of $a_{\mathsf{S}}$ in Fig.\ \ref{mae_vs_a}(c), holding $a_{\mathsf{T}}=1$. The main feature here is the symmetry of all the performance curves: around $a_{\mathsf{S}}=0$ for FPDa, ICM and SNT; around $a_{\mathsf{S}}=1$ for FPDb; and (trivially) everywhere for TNT. In all cases, this is due to the fact that $a_{\mathsf{S}}$ quadratically enters the correlation structure, $B$ \eqref{coregionalization_matrix}, of the global target modeller. When $\sigma_{\mathsf{S}}^{2}\neq\sigma_{\mathsf{T}}^{2}$ (e.g.\ the case in Fig.\ \ref{mae_vs_a}(d) where $\sigma_{\mathsf{S}}^{2}\ll\sigma_{\mathsf{T}}^{2}$), the main point to note is that FPDb cannot benefit from the improved positive transfer available from this more precise source. FPDa does benefit from this source precision because it, uniquely, exploits the target's correlation structure,~$B$.

In summary, Fig.\ \ref{mae_vs_a} illustrates the fact that the ICM algorithm---with exactly matched synthesis and analysis models---delimits the best predictive performance that can be attained by the FPDa algorithm. That said, misspecification of the ICM analysis model---as will occur with probability one in real-data settings---will undermine its performance, and provide opportunity for FPDa to outperform it. This is so, since FPDa allows source and target to elicit their models independently (Fig. \ref{graphical_model}(c)). We demonstrate this crucial model robustness feature of our BTL algorithm in the next section.

\subsection{Transfer of the source's (local) analysis model (Fig.\ \ref{structure})}\label{source_model_structure_section}
In our BTL framework (i.e.\ FPDa), the source modeller independently chooses a different model from the global target modeller to drive its source learning task. The source distribution, $\mathsf{F}_{\mathsf{S}}$ \eqref{posterior_predictor}, therefore encodes not only the information from the local source data, $\mathbf{y}_{\mathsf{S}}$, but also the source model structure. In this section, we demonstrate the features of this approach under conditions of mismatch (i.e.\ disagreement) between the source's and target's model of $\mathbf{y}_{\mathsf{S}}$. Specifically, we will now design an experiment where the source modeller more closely captures the synthesis model than the global target modeller does; i.e.\ the source exhibits \emph{local expertise}, as is often the case in practice.

\textit{Settings}:\ The purpose of this study is to explore the impact of kernel structure mismatch between the global target's analysis model and the ICM synthesis model. In all these cases, we arrange for the source analysis model to match the synthesis model, achieving local source expertise (above). The structures of the synthesis and analysis models are varied only via the kernel function, $k_{\theta_{u}}(\cdot,\cdot)$ \eqref{pre_prior_covariance_matrix_specific}, and its parameters, $\theta_{u}$, for the seven cases in Table~\ref{kernels}. The specific choices of these parameters in our current study are as follows: $(\sigma^{2},\gamma,d,l^{2},\alpha)\equiv(1.0,1.0,3,0.2,1.0)$. Recall that---in the previous simulation study (Fig.\ \ref{mae_vs_a})---the remaining parameters, $(\sigma_{\mathsf{S}}, \sigma_{\mathsf{T}},a_{\mathsf{S}}, a_{\mathsf{T}})$, influence the transfer learning properties of the ICM and FPDa algorithms. We choose the settings, $\sigma_{\mathsf{S}}=\sigma_{\mathsf{T}}=1$ and $a_{\mathsf{S}}=a_{\mathsf{T}}=1$, where all the transfer algorithms benefit equally from the source data, and so the only differential benefit between the algorithms will be in respect of the quality of the analysis model adopted by the source in computing its transferred knowledge.

\textit{Performance}:\ Fig.\ \ref{structure} shows the MAE for all possible combinations of the kernel functions in the ICM synthesis model (Table \ref{kernels}), and, again adopted in the TNT, ICM, and FPDa analysis models of the respective algorithms in Table~\ref{algorithms} (i.e.\ these are $7\times 7 = 49$ kernel combinations \emph{in toto}). Since the SNT algorithm is arranged to capture perfectly the source part of the ICM synthesis model, therefore it requires only a single column in Fig.\ \ref{structure}(a). We do not present the FPDb algorithm, since---with the currently adopted experimental settings---it reduces to the FPDa algorithm (Fig.\ \ref{mae_vs_a}(c)). The TNT algorithm delineates the baseline MAE performance of no transfer (as it did in Section \ref{performance_bound_section}), allowing us to assess the extent if positive transfer for the various kernel settings in ICM and FPDa. Note that both algorithms stay in the positive transfer regime for all combinations of the synthesis and analysis models, thus benefiting from the source knowledge even in the misspecification cases. For this reason, we have illustrated the absolute value differential MAE in Fig.\ \ref{structure}(c) and (d) (i.e.\ higher values are better in these two sub-figures). Furthermore, the red asterisks indicate those kernel combinations for which our FPDa algorithm outperforms the ICM algorithm. Fig.\ \ref{structure}(c) and (d) reveal that the ICM algorithm slightly outperforms the FPDa algorithm on all diagonal entries, i.e.\ where there is no mismatch between the synthesis and analysis models. This is consistent with our finding in Section \ref{performance_bound_section}, namely that the ICM algorithm achieves the optimal performance in this (unrealistic) case of perfect analysis-synthesis model matching. Conversely, the FPDa algorithm demonstrates improved robustness to kernel structure misspecification when compared to the ICM algorithm, particularly when FPDa adopts a more complex analysis model---i.e.\ SE, QR, M (Table \ref{kernels})---and the synthesis model is simple, i.e.\ C, L, P, CO. This can be seen from the upper-right yellow quadrant in Fig.\ \ref{structure}(d). These more complex kernel functions for analysis are, indeed, a common choice for real data, where it is often hard to choose a kernel function based on a mere visual inspection only. The remaining analysis-synthesis combinations---especially those under the diagonal in Fig.\ \ref{structure}(d)---yield FPDa predictive performances in the target that are almost indistinguishable from those of the ICM algorithm.%

\begin{figure*}[t]
	\includegraphics{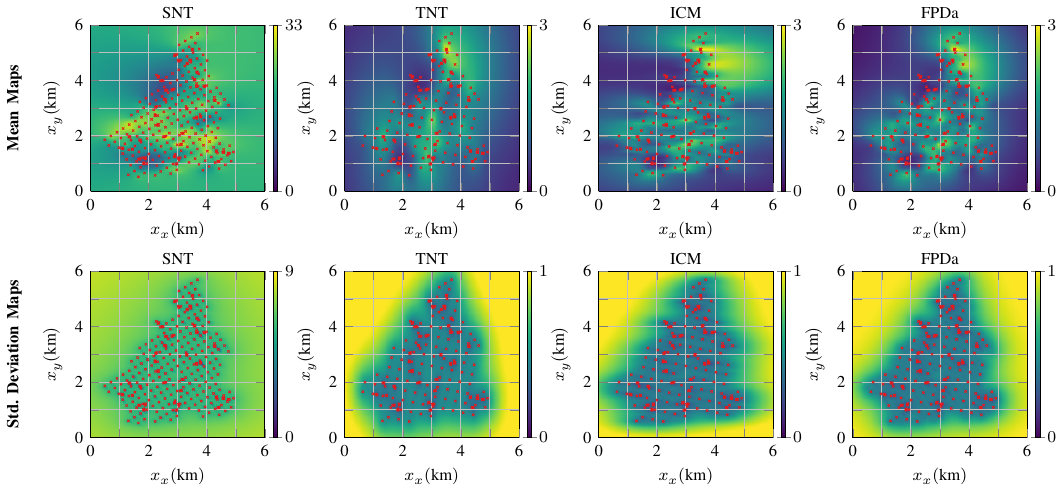}
	\vspace{-5pt}
	\caption{
		\textit{Real-data experiments}. Marginal concentration predictions of Ni (ppm) via the SNT algorithm, and of Cd via the TNT, ICM, and FPDa algorithms (Table~\ref{algorithms}) in a region of the Swiss Jura. \textbf{Top}: the marginal predictive mean map at super-sampled spatial coordinates, $x^{\ast}\equiv(x^{\ast}_{x},x^{\ast}_{y})$; \textbf{bottom}: the associated marginal predictive standard deviation maps. The red crosses are the locations of observed Ni and Cd concentrations used for training (359 for Ni in the source task (first column), and 259 for Cd in the target task (remaining columns)).
	}
	\label{prediction_performance}
\end{figure*}

\section{Real-Data Experiments}\label{real_data_experiments}
The Swiss Jura geological dataset\footnote{https://sites.google.com/site/goovaertspierre/} records the concentrations of seven heavy metals---measured in units of parts-per-million (ppm)---at $359$ (spatial) locations in a 14.5 $\text{km}^2$ region. This dataset has been widely studied in the context of multitask learning (see, for example, \cite{alvarez2011computationally}). In our study, the objective is to predict concentrations of the less easily detected metal, cadmium (Cd), based on knowledge of the more easily detected metal, nickel (Ni). In order to demonstrate our BTL approach (Algorithm \ref{FPD}) in this context, the source learning task Ni concentrations, and transfers its probabilistic output-data predictor to the target learning task of Cd concentration prediction. Note that we perform transfer with only this Ni source task, whereas, in \cite{alvarez2011computationally}, zinc (Zn) concentrations were also processed at the source.

\textit{Alignment with the FPD framework}:\ The SNT algorithm processes the Ni concentrations measured at all $n_{\mathsf{S}}=359$ locations (i.e.\ source data, $\mathbf{y}_{\mathsf{S}}$) and uses these to compute the Ni output-data predictor~\eqref{posterior_predictor}. The FPDa algorithm processes $n_{\mathsf{T}}=259$ of the measured Cd concentrations (i.e.\ target data, $\mathbf{y}_{\mathsf{T}}$), along with the transferred Bayesian predictive moments, $\bar{\mathbf{m}}_{\mathsf{S}}$ and $\bar{\mathbf{R}}_{\mathsf{S}}$, of the source Ni output-data predictor~\eqref{posterior_predictor}, in order to improve the prediction of the Cd concentrations at the  $n^{\ast}=100$ hold-out locations, $\mathbf{x}^{\ast}\equiv(x^{\ast}_{i})^{n^{\ast}}_{i=1}$, with spatial coordinates $x^{\ast}_{i}\equiv(x_{x,i}^{\ast},x_{y,i}^{\ast})$. The aim is to compute $\mathbf{m}_{\mathsf{T}}^{\mathsf{o}\ast}\equiv m_{\mathsf{T}}^{\mathsf{o}}(\mathbf{x}^{\ast})$ and $\mathbf{K}_{\mathsf{TT}}^{\mathsf{o}\ast}\equiv k_{\mathsf{TT}}^{\mathsf{o}}(\mathbf{x}^{\ast},\mathbf{x}^{\ast})$, where $m_{\mathsf{T}}^{\mathsf{o}}$ and $k_{\mathsf{TT}}^{\mathsf{o}}$ are the lower sub-vector and lower-right sub-matrix of \eqref{optimal_posterior_gp_mean_function} and \eqref{optimal_posterior_gp_covariance_function}, respectively. In the Ni source task (SNT), the MAE for the GP predictive mean \eqref{source_posterior_mean_function} was computed across a wide range of candidate kernel functions, weighting the inputs, $\mathbf{x}_{\mathsf{S}}$, via automatic relevance determination (ARD) \cite{rasmussen2006gaussian}. The optimal choice was the rational quadratic kernel (Table \ref{kernels}). In the isolated Cd target task (also with ARD), the Mat\'{e}rn kernel proved to be optimal, and was adopted for all the target learning algorithms (Table \ref{algorithms}).

\textit{Parameter learning}:\ The notion of a synthesis model (of the type in \eqref{synthesis_model}) does not, of course, arise in this real-data context. Instead, we adopt maximum likelihood (ML) estimation \cite{rasmussen2006gaussian} in order to learn the parameters of the analysis models underlying the various algorithms in Table \ref{algorithms}. For this purpose, we use the iterative L-BFGS-B algorithm \cite{zhu1997algorithm}---widely available as a library tool---to compute a local optimum of the log-marginal likelihood surface. The maximum number of iterations is limited to 20,000. We initialize the parameters (Table \ref{kernels}) of the rational quadratic kernel with $(\sigma^{2}_{u},l^{2}_{u},\alpha_{u})\equiv(1.0,1.0,1.0)$ and the parameters of the Mat\'{e}rn kernel with $(\sigma^{2}_{u},l^{2}_{u})\equiv(1.0,1.0)$, where the value of the length-scale, $l^{2}_{u}$, now applies to both input dimensions of $x$ in accordance with the ARD mechanism. The coefficients, $a_{\mathsf{S}}$ and $a_{\mathsf{T}}$, parameterizing the coregionalization matrix, $B$ \eqref{coregionalization_matrix}, are initialized as $(a_{\mathsf{S}}, a_{\mathsf{T}})=(0.1,0.1)$. For this Jura dataset, we found that the adopted ML procedure was insensitive to perturbations of these initial parameters.

\begin{table}[t]
	\centering
	\footnotesize
	\begin{tabular}{c|c}
		\hline
		Algorithm & MAE \\
		\hline
		TNT  & 5.9273e$-$01 $\pm$ 0.0001e$-$01\\
		ICM  & 5.2808e$-$01 $\pm$ 0.0003e$-$01\\
		FPDa & 4.9966e$-$01 $\pm$ 0.0005e$-$01\\
		FPDb & 1.9510e$+$01 $\pm$ 0.0003e$+$01\\
		\hline
	\end{tabular}
	\caption{\textit{Real-data experiments}. The MAE \eqref{mae} of predicted Cadmium concentrations for the TNT, ICM, FPDa, and FPDb algorithms in the Jura dataset. The results are averaged over 10 runs with different initialization of the estimated unknown parameters, presenting the mean and standard deviation.}
	\label{jura_results}
\end{table}

\textit{Performance}:\ To assess quantitatively the prediction performance of the algorithms in Table \ref{algorithms}, we compute the MAE \eqref{mae} across the $n^{\ast}=100$ hold-out locations where we wish to predict Cadmium. We find that the FPDa algorithm delivers positive transfer (i.e.\ it has a lower MAE than the baseline TNT algorithm), and, importantly, it outperforms the ICM algorithm. This improves on our former FPDb algorithm which suffers negative transfer because---as explained in Section \ref{synthetic_data_experiments}---it is not equipped to adopt and learn the target's correlation structure via $B$ \eqref{coregionalization_matrix}. Our FPDa performance is close to that reported in \cite{alvarez2011computationally}, despite the fact that we transfer predictive source knowledge only from Ni concentrations, whereas Zn concentrations are also processed in \cite{alvarez2011computationally}, as already noted above. Returning to our own FPDa algorithm and the alternatives in Table \ref{algorithms}, we summarize our findings in Fig.\ \ref{prediction_performance}. In the top row, we illustrate the spatial maps of the marginal predictive mean concentration of Ni (SNT) and Cd (TNT, ICM, FPDa), equipping these with uncertainty represented by the corresponding marginal predictive standard deviation maps (bottom row). These have been computed on a uniform $50\times 50$ spatial grid of the input, $x^{\ast}\equiv(x^{\ast}_{x},x^{\ast}_{y})$. In this sense, the current study can be characterized as an image reconstruction task, driven by non-uniformly sampled and noisy measurements (i.e.\ pixel values), with registration between the source and target images. The Mat\'{e}rn kernel provides the required prior regularization that induces spatial smoothness in the (Cd) reconstruction \cite{ring1994continual,ton2018spatial}. Note that the FPDb algorithm is not illustrated in Fig.\ \ref{prediction_performance}, since it performs poorly (Table~\ref{jura_results}).

Visual inspection of the TNT, ICM and FPDa columns of Fig.\ \ref{prediction_performance} reveals that our algorithm (FPDa) localizes the Cd deposits more sharply than either the ICM algorithm, or (unsurprisingly) the TNT algorithm, which does not benefit from any Ni source learning. In particular, note the horizontal blurring of the Cd predictive mean map for ICM (third column, top), an artefact that is avoided in FPDa (fourth column, top). This localization of the FPDa prediction supports the exploration of Cd deposits in a focussed area around $x^{\ast}=(3.8, 4.6)$ $\text{km}^{2}$. In contrast, the ICM predictive mean map does not sufficiently resolve the $x_{x}$ coordinate to support on exploration decision in this area.

\section{Discussion}\label{discussion}
We will now assess our FPD-optimal BTL framework (i.e.\ Algorithm \ref{FPD} (FPDa)) in the context of the conventional Bayesian multitask learning framework (via the ICM algorithm). We will focus on a number of specific themes that emerge from the evidence in Sections \ref{synthetic_data_experiments} and \ref{real_data_experiments}.

\textit{Transfer of moments:}\ In the previously considered Gaussian settings \cite{foley2018fully,papez2018dynamic,papez2019bayesian} of our FPD-optimal BTL framework, the second-order moments of the output-data predictor, $\mathsf{F}_{\mathsf{S}}$, were not successfully transferred, leading to negative transfer. This is also true of FPDb in this paper (Fig.\ \ref{mae_vs_a}(a) and (b) and Table \ref{jura_results}). In all those papers, the order of arguments in the KLD was the reverse of the one proposed in this paper \eqref{fpd_optimal_posterior_distribution_generic}, a reversal which is essential for the robust transfer property of our FPDa algorithm. This reversal has ensured the successful transfer of the second-order moment, $\bar{\mathbf{R}}_{\mathsf{S}}$, of \eqref{posterior_predictor}, even in the current Gaussian setting, yielding the block-diagonal covariance structure \eqref{optimal_posterior_gp_functions}. Equipping the FPDa algorithm with $\bar{\mathbf{R}}_{\mathsf{S}}$ delivers robust knowledge transfer, as demonstrated in the following remark. 
\begin{rem}\label{remark_robustness}
	Consider \eqref{optimal_posterior_gp_functions} for $q=p=\mathsf{T}$ (i.e.\ the target after transfer) in the limit where the source knowledge becomes uniform (and, therefore, non-informative \cite{jeffreys1961theory}). In this case,
	\begin{align}
		m_{\mathsf{T}}^{\mathsf{o}}(x)&\rightarrow \bar{m}(x),
		\nonumber
		\\
		k_{\mathsf{TT}}^{\mathsf{o}}(x,x')&\rightarrow \bar{k}(x,x'),
		\nonumber
	\end{align}
	being the moments of the isolated TNT learning task. This rejection of non-informative source knowledge by the target in our FPDa algorithm is what we mean by robust transfer.
\end{rem}
The role of $\bar{\mathbf{R}}_{\mathsf{S}}$ in effecting robust transfer---as well as in strengthening the positive transfer above threshold which we observed in Fig.\ \ref{mae_vs_a}(a) and (b)---will be studied more technically in a forthcoming paper.

\textit{Weighting of the transferred knowledge:}\ Recall that the matrix, $\mathbf{K}$ \eqref{pre_prior_covariance_matrix}---which is instantiated in the ICM case in \eqref{pre_prior_covariance_matrix_specific} and \eqref{coregionalization_matrix}---expresses the target's interaction with the source. Inspecting \eqref{optimal_posterior_gp_functions} and \eqref{interaction_structure}, we also see that $\mathbf{K}$ controls the weighting mechanism in the FPDa algorithm, tuning the influence of the transferred knowledge, $\mathsf{F}_{\mathsf{S}}$, on the optimal target conditional, $\mathsf{M}^{\mathsf{o}}(\mathbf{f}^{}_{\mathsf{T}}|\mathsf{F}_{\mathsf{S}},\mathbf{y}^{}_{\mathsf{T}},\mathbf{K},\cdot)$ \eqref{fpd_optimal_posterior_distribution_generic}. This is seen in the fact that its target moments, $\mathbf{m}^{\mathsf{o}}_{\mathsf{T}}$ and $\mathbf{K}^{\mathsf{o}}_{\mathsf{TT}}$, are functions of $\mathbf{K}$. The study of this behaviour can be formalized, by adopting similar limit arguments as in Remark \ref{remark_robustness}. Furthermore, the optimization of $\mathbf{K}$---which is a hyperparameter of $\mathsf{M}^{\mathsf{o}}(\mathbf{f}^{}_{\mathsf{T}}|\mathsf{F}_{\mathsf{S}},\mathbf{y}^{}_{\mathsf{T}},\mathbf{K},\cdot)$ (see \eqref{fpd_optimal_posterior_distribution_specific})---via maximum likelihood estimation (Section \ref{real_data_experiments}) ensures that this weighting is data-driven (i.e.\ adaptive). This is also an advance beyond our previous proposals for FPD-optimal BTL \cite{foley2018fully,papez2018dynamic,papez2019bayesian}, in which no transfer weighting mechanism was induced. This progress has been achieved only because the target modeller is an extended modeller of the interacting source-target tasks, unlike the situation in our previous work where the target models only its local task. Technical details relating to this inducing of a data-driven transfer weighting will be developed further in future work.

\textit{Robustness to analysis-synthesis model mismatch:}\
In simulations contexts---such as in Section \ref{synthetic_data_experiments}---the data are simulated from the chosen \emph{synthesis model}, while each assessed algorithm is consistent with an \emph{analysis model}. This provides an opportunity to assess the robustness of each algorithm to model misspecification, as studied in Section \ref{source_model_structure_section}. We have already noted that the ICM algorithm achieved a performance optimum across all the studied algorithms in Section \ref{synthetic_data_experiments} since its (ICM) analysis model equals the (ICM) synthesis model used to generate all data in that study. Nevertheless, the FPDa algorithm performs almost as well (Fig.\ \ref{mae_vs_a}), despite the fact that it does not rely on complete instantiation of the synthesis model. Indeed, as explained in Section \ref{btl_for_global_modellers}, FPD-optimal BTL does not require completion of the interaction model between the source and target at all, and has therefore proved to be more robust to ignorance of the synthesis model than the ICM algorithm has. In real-data applications---such as in Section \ref{real_data_experiments}---there is, of course, no synthesis model. The benefit of our incompletely modelled approach in terms of robustness has been demonstrated: FPDa can provide a closer fit to the data than ICM, as presented in Fig. \ref{prediction_performance} and Table \ref{jura_results}. The SNT task involves a second independent local modeller (Fig.\ \ref{graphical_model}(c)) which constructs its source output-data predictor, $\mathsf{F}_{\mathsf{S}}$ \eqref{posterior_predictor}, using the same source data that the ICM algorithm processes. The source's local modelling expertise (Section \ref{source_model_structure_section}) has enriched the knowledge transfer, and provided a supplementary learning resource which is not available to ICM.

\textit{Computational load of our BTL algorithm:}\ The principal computational cost of all the algorithms in Table \ref{algorithms} resides in inverting the linear systems involved in the standard Gaussian (conditional) data update \eqref{posterior_gp_functions}. This cost scales cubically with the number of data in the source and target tasks, respectively, i.e.\ $\mathcal{O}(n_{q}^{3})$, $q\in(\mathsf{S},\mathsf{T})$ \cite{liu2020gaussian}. Owing to the sequential nature of the knowledge processing in FPDb (i.e.\ source processing of raw data $\mathbf{y}_{\mathsf{S}}$, \emph{then} target processing of the transferred source Bayesian predictor, $\mathsf{F}_{\mathsf{S}}$, along with the target raw data, $\mathbf{y}_{\mathsf{T}}$), the net computational load of FPDb is $\mathcal{O}(n_{\mathsf{S}}^{3})+\mathcal{O}(n_{\mathsf{T}}^{3})$. In contrast, our new BTL algorithm (FPDa) shares the property with ICM of bidirectional knowledge flow (Fig.\ \ref{graphical_model}(c)). This leads to the matrix augmentation form of the inverse systems in \eqref{optimal_posterior_gp_functions} and, correspondingly, a more computationally expensive---$\mathcal{O}((n_{\mathsf{S}}+n_{\mathsf{T}})^{3})$---algorithm. There is a rich literature on efficient inversions for GP regression, exploiting possible reductions (notably sparsity) in the covariance matrix, $\mathbf{K}$ \eqref{pre_prior_covariance_matrix}. These same reductions have been applied in multitask GP learning \cite{liu2020gaussian}, such as in ICM. Since our new FPDa BTL algorithm shares the same augmented matrix structure \eqref{optimal_posterior_gp_functions} as ICM, it, too, can benefit from these reductions.

\textit{Source knowledge compression:}\ Conventional multitask learning---such as in Fig.\ \ref{graphical_model}(b)---requires the communication of all $n_{\mathsf{S}}$ of the source's extended (training) data, $(\mathbf{x}_{\mathsf{S}}, \mathbf{y}_{\mathsf{S}})$, to the processor (either centrally or in some distributed manner), i.e.\ the transferred message size (quantified as the number of scalars) is $n_{\mathsf{S}}(n_{x_{\mathsf{S}}} + 1)$. Conversely, our FPDa algorithm (Fig.\ \ref{graphical_model}(c)) is a truly \emph{Bayesian} transfer learning algorithm, whose communication overhead resides only in the transfer of the source's sufficient statistics, $(\bar{\mathbf{m}}_{\mathsf{S}},\bar{\mathbf{R}}_{\mathsf{S}})$, of its output-data predictor at predictive inputs, $\mathbf{x}^{\star}_{\mathsf{S}}$, i.e.\ $\mathsf{F}_{\mathsf{S}}(\mathbf{y}_{\mathsf{S}}^{\star}|\mathbf{y}_{\mathsf{S}},\mathbf{x}_{\mathsf{S}},\mathbf{x}^{\star}_{\mathsf{S}},\theta_{\mathsf{S}})=\mathcal{N}(\bar{\mathbf{m}}_{\mathsf{S}},\bar{\mathbf{R}}_{\mathsf{S}})$ \eqref{posterior_predictor}. Therefore, the transferred knowledge is fully encoded by these source sufficient statistics without the requirement to transfer the raw extended source data, $(\mathbf{y}_{\mathsf{S}},\mathbf{x}^{}_{\mathsf{S}},\mathbf{x}^{\star}_{\mathsf{S}})$. The message size for our BTL algorithm is therefore $n^{\star}_{\mathsf{S}}(1+\tfrac{n^{\star}_{\mathsf{S}}+1}{2})$. It follows that the condition which must hold for compression to be achieved in our BTL algorithm versus conventional raw data transfer is\footnote{Here, we quote the implied inequality for the special case, $n^{\star}_{\mathsf{S}}\equiv n_{\mathsf{S}}$.} $n^{\star}_{\mathsf{S}}<2n_{x_{\mathsf{S}}}-1$, i.e.\ the number of predictive points must be lower than two times the dimension of each input (minus one). This condition expresses the objective of avoiding dilution of source knowledge in the predictor \cite{wainwright2019high}. Note also that the source sufficient statistics are functions of the source GP kernel structure, $k_{\mathsf{S}}$ (Remark \ref{source_gp_regression_remark}). This can be highly expressive---with many degrees of freedom---in cases of local source expertise. We have provided a preliminary study of the benefit to the target of a distinct source covariance structure (Section \ref{source_model_structure_section}, and see also the third discussion theme above). Indeed, far more expressive covariance structures in $k_{\mathsf{S}}$---involving weighted combinations of canonical kernel functions (Table \ref{kernels})---can also be adopted by the source \cite{parra2017spectral}. In all cases, our Bayesian knowledge transfer compresses this resource into dimension-invariant sufficient statistics for subsequent processing by the target~task.

\vspace{-4pt}
\begin{center}
---------
\end{center}
We now summarize key findings in respect of our BTL framework (and its incomplete modelling of interaction between the independent source and target modellers (Fig.\ \ref{graphical_model}(c))) versus conventional multitask learning (and its complete and unitary modelling of all source-target variables (Fig.\ \ref{graphical_model}(b))).

\begin{itemize}[leftmargin=12pt]
	\item Our FPD-optimal BTL algorithm (Fig.\ \ref{graphical_model}(c)) is truly a \emph{transfer} learning algorithm---as opposed to a multitask learning algorithm---because it does not require complete modelling of source-target interactions. Among the consequences we discovered in this paper are (i) the possibility to avoid problems of model misspecification, and (ii) the opportunity to transfer potentially compressed source sufficient statistics of the data-predictive distribution, $\mathsf{F}_{\mathsf{S}}$, instead of the raw data themselves.

	\item The source and target modellers (Fig.\ \ref{graphical_model}(c)) are independent, and, indeed, our BTL algorithm is truly a multiple model algorithm, in contrast to the single global modeller of multitask learning (Fig.\ \ref{graphical_model}(b)). We have shown that a direct consequence is the ability of our algorithm to transfer source knowledge enriched by an expressive local kernel structure.

	\item Our BTL algorithm delegates the processing of local data to the local (source) task before the resulting stochastic message, $\mathsf{F}_{\mathsf{S}}$, is transferred for subsequent processing by the target task (along with its local (target) data). This opens up the possibility for distributed and parallel processing in a way that is not intrinsic to multitask learning.

	\item In our formulation of BTL (Fig.\ \ref{graphical_model}(c)), we have exploited the opportunity to transfer knowledge to a target task that is, itself, a global modeller. The paper has presented evidence to show that this extension of the target ensures that unreliable source knowledge is rejected (i.e.\ robust transfer is achieved (Remark \ref{remark_robustness})), and our positive transfer above threshold is as good as that of conventional multitask learning, such as ICM.
\end{itemize}

\section{Conclusion}\label{conclusion}
A vulnerability of conventional multitask learning arises from the fact that there must exist a global modeller of all the tasks in the system. This imposes a requirement on the global modeller to capture task interactions accurately, and performance is undermined otherwise. Furthermore, it is intrinsic to that framework that the global modeller must process raw data from the remote sources. This potentially incurs a communication overhead and forces the target to process remote source data about which it may lack local expertise. Against this, the complete nature of multitask modelling ensures its optimal performance in the (admittedly unlikely) case where task interactions are accurately modelled.

The FPD-optimal Bayesian transfer learning (BTL) framework developed and tested in this paper has achieved important progress beyond the conventional state-of-the-art above. Its key advance is that it does not require elicitation of a model of dependence between the interacting tasks (a defining aspect of BTL, in our opinion, called \emph{incomplete modelling}), and so our approach avoids the misspecification that inevitably arises in conventional multitask learning. Instead, it chooses the conditional target distribution---from among the uncountable  cases consistent with the partial model and with the knowledge constraints---in a decision-theoretically optimal manner.
A number of important practical benefits flow from this. Firstly, the local source processes its local data---exploiting its local modelling expertise---into a Bayesian (i.e.\ probabilistic) predictor. Secondly, only the sufficient statistics of this predictor need be transferred to the target. Thirdly, the target \emph{then} sequentially has the opportunity to process this source knowledge along with its local target data. Fourthly, our framework allows this target to be a global modeller of both tasks, ensuring that optimal positive transfer is preserved in the case specified at the end of the previous paragraph.

The success of our algorithm hinges on its ability to transfer all moments of the source predictor, an advance beyond our earlier variants of BTL, and achieved via careful specification of the KLD objective \eqref{fpd_optimal_posterior_distribution_generic}. The source covariance matrix, $\bar{\mathbf{R}}_{\mathsf{S}}$, has been vital in knowledge-driving the weighting attached to the transferred predictor by the target. This has obviated the need for hierarchical relaxation of the target model which was necessary in our previous work in order to learn this weight, and which incurred relatively expensive variational approximation \cite{papez2019robust,papez2020bayesian}.

The FPD-optimal BTL design of the target's source-conditional learning \eqref{fpd_optimal_posterior_distribution_generic} can be interpreted as a binary operator, closed within the class of probability distributions. It operates (non-commutatively) on the transferred source output-data predictor, $\mathsf{F}_{\mathsf{S}}$, and on the target's posterior distribution, $\mathsf{F}(\mathbf{f}^{}_{\mathsf{S}},\mathbf{f}^{}_{\mathsf{T}}|\mathbf{y}^{}_{\mathsf{T}})$, yielding the optimal-knowledge-conditional target posterior inference, $\mathsf{M}^{\mathsf{o}}(\mathbf{f}^{}_{\mathsf{S}},\mathbf{f}^{}_{\mathsf{T}}|\mathsf{F}_{\mathsf{S}},\mathbf{y}^{}_{\mathsf{T}})$. This closure of our BTL operator within the class of probability distributions recommends it in a wide range of continual learning tasks \cite{ring1994continual,nguyen2018variational}. For instance, in incompletely modelled networks, the requirement for all BTL knowledge objects to be distributions ensures that local computational resources onboard local nodes are exploited in processing local data into these local distributions. Furthermore, target nodes can recursively act as source nodes in a continual learning process, so that probabilistic decision-making is distributed across a network, subject to specification of the network architecture. Such applications of our work in Bayesian networks---and, indeed, in deep learning contexts \cite{lecun2015deep}---can provide a rich opportunity for our Bayesian transfer learning paradigm to address key technology challenges at this time.

\appendices
\section{Proofs}
\subsection{Preliminaries}
\begin{lem}\label{gaussian_joint_density}
	Let us consider the following factorized, jointly Gaussian model:
	\begin{align}
		\mathsf{F}(\mathbf{y}|\mathbf{f})
		&\equiv
		\mathcal{N}(\mathbf{y};\mathbf{f},\mathbf{\Sigma})
		,
		\nonumber
		\\
		\mathsf{F}(\mathbf{f})
		&\equiv
		\mathcal{N}(\mathbf{f};\mathbf{m},\mathbf{K})
		.
		\nonumber
	\end{align}
	Then, the joint distribution is
	\begin{equation}
		\mathsf{F}(\mathbf{f},\mathbf{y})
		=
		\mathcal{N}
		\bigg(
			\left[
				\begin{matrix}
					\mathbf{f}
					\\
					\mathbf{y}
				\end{matrix}
			\right]
			;
			\left[
				\begin{matrix}
					\mathbf{m}
					\\
					\mathbf{m}
				\end{matrix}
			\right]
			,
			\left[
				\setlength\arraycolsep{2pt}
				\begin{matrix}
					\mathbf{K} & \mathbf{K}
					\\
					\mathbf{K} & \mathbf{K}+\mathbf{\Sigma}
				\end{matrix}
			\right]
		\bigg)
		.
		\label{joint_gaussian}
	\end{equation}
	\begin{proof}
		The proof follows from standard analysis.
	\end{proof}
\end{lem}

\begin{lem}\label{gaussian_conditional_and_marginal_densities}
	Let us consider a slight generalization of the Gaussian distribution in \eqref{joint_gaussian}:
	\begin{equation}
		\mathsf{F}(\mathbf{f},\mathbf{y})
		\equiv
		\nonumber
		\mathcal{N}
		\bigg(
			\left[
				\begin{matrix}
					\mathbf{f}
					\\
					\mathbf{y}
				\end{matrix}
			\right]
			;
			\left[
				\begin{matrix}
					\bar{\mathbf{f}}
					\\
					\bar{\mathbf{y}}
				\end{matrix}
			\right]
			,
			\left[
				\setlength\arraycolsep{2pt}
				\begin{matrix}
					\mathbf{K} & \mathbf{K}
					\\
					\mathbf{K} & \mathbf{K}+\mathbf{\Sigma}
				\end{matrix}
			\right]
		\bigg)
		.
		\nonumber
	\end{equation}
	Then, the conditional and marginal densities are
	\begin{align}
		\mathsf{F}(\mathbf{f}|\mathbf{y})
		&=
		\mathcal{N}(\mathbf{f};\bar{\mathbf{m}},\bar{\mathbf{K}})
		,
		\nonumber
		\\
		\mathsf{F}(\mathbf{y})
		&=
		\mathcal{N}(\mathbf{y};\bar{\mathbf{y}},\mathbf{K}+\mathbf{\Sigma})
		,
		\nonumber
	\end{align}
	where
	\begin{align}
		\bar{\mathbf{m}}
		&=
		\bar{\mathbf{f}}+\mathbf{K}(\mathbf{K}+\mathbf{\Sigma})^{-1}(\mathbf{y}-\bar{\mathbf{y}})
		,
		\nonumber
		\\
		\bar{\mathbf{K}}
		&=
		\mathbf{K}-\mathbf{K}(\mathbf{K}+\mathbf{\Sigma})^{-1}\mathbf{K}
		.
		\nonumber
	\end{align}
	\begin{proof}
		The proof follows from standard analysis.
	\end{proof}
\end{lem}

\subsection{Proof of Proposition \ref{fpd_optimal_static_transfer}}\label{fpd_optimal_static_transfer_proof}
By substituting the constrained unknown and ideal models, \eqref{restricted_unknown_joint_model} and \eqref{ideal_model} respectively, into \eqref{fpd_optimization_problem}, we can write
\begin{align}
	\mathcal{D}(\mathsf{M}||\mathsf{M}_{\mathsf{I}})
	\nonumber
	\\
	&\hspace{-40pt}=
	\int\mathsf{F}(\mathbf{y}_{\mathsf{S}}^{\star}|\mathbf{f}^{}_{\mathsf{S}},\theta)\mathsf{M}(\mathbf{f}^{}_{\mathsf{S}},\mathbf{f}^{}_{\mathsf{T}}|\mathsf{F}_{\mathsf{S}},\mathbf{y}^{}_{\mathsf{T}},\theta)
	\nonumber
	\\
	&\hspace{-30pt}\times
	\log\frac{\mathsf{F}(\mathbf{y}_{\mathsf{S}}^{\star}|\mathbf{f}^{}_{\mathsf{S}},\theta)\mathsf{M}(\mathbf{f}^{}_{\mathsf{S}},\mathbf{f}^{}_{\mathsf{T}}|\mathsf{F}_{\mathsf{S}},\mathbf{y}^{}_{\mathsf{T}},\theta)}{\mathsf{F}_{\mathsf{S}}(\mathbf{y}_{\mathsf{S}}^{\star}|\mathbf{y}^{}_{\mathsf{S}})\mathsf{F}(\mathbf{f}^{}_{\mathsf{S}},\mathbf{f}^{}_{\mathsf{T}}|\mathbf{y}^{}_{\mathsf{T}},\theta)}d\mathbf{y}_{\mathsf{S}}^{\star}d\mathbf{f}^{}_{\mathsf{S}}d\mathbf{f}^{}_{\mathsf{T}}
	\nonumber
	\\
	&\hspace{-40pt}=
	\int\mathsf{M}(\mathbf{f}^{}_{\mathsf{S}},\mathbf{f}^{}_{\mathsf{T}}|\mathsf{F}_{\mathsf{S}},\mathbf{y}^{}_{\mathsf{T}},\theta)\log\bigg[\frac{\mathsf{M}(\mathbf{f}^{}_{\mathsf{S}},\mathbf{f}^{}_{\mathsf{T}}|\mathsf{F}_{\mathsf{S}},\mathbf{y}^{}_{\mathsf{T}},\theta)}{\mathsf{F}(\mathbf{f}^{}_{\mathsf{S}},\mathbf{f}^{}_{\mathsf{T}}|\mathbf{y}^{}_{\mathsf{T}},\theta)}
	\nonumber
	\\
	&\hspace{-30pt}+
	\int \mathsf{F}(\mathbf{y}_{\mathsf{S}}^{\star}|\mathbf{f}^{}_{\mathsf{S}},\theta)\log\frac{\mathsf{F}(\mathbf{y}_{\mathsf{S}}^{\star}|\mathbf{f}^{}_{\mathsf{S}},\theta)}{\mathsf{F}_{\mathsf{S}}(\mathbf{y}_{\mathsf{S}}^{\star}|\mathbf{y}^{}_{\mathsf{S}})}d\mathbf{y}_{\mathsf{S}}^{\star}\bigg]d\mathbf{f}^{}_{\mathsf{S}}d\mathbf{f}^{}_{\mathsf{T}}
	\nonumber
	\\
	&\hspace{-40pt}=
	\int\mathsf{M}(\mathbf{f}^{}_{\mathsf{S}},\mathbf{f}^{}_{\mathsf{T}}|\mathsf{F}_{\mathsf{S}},\mathbf{y}^{}_{\mathsf{T}},\theta)
	\nonumber
	\\
	&\hspace{-30pt}\times
	\log\frac{\mathsf{M}(\mathbf{f}^{}_{\mathsf{S}},\mathbf{f}^{}_{\mathsf{T}}|\mathsf{F}_{\mathsf{S}},\mathbf{y}^{}_{\mathsf{T}},\theta)}{\mathsf{F}(\mathbf{f}^{}_{\mathsf{S}},\mathbf{f}^{}_{\mathsf{T}}|\mathbf{y}^{}_{\mathsf{T}},\theta)\exp\left\lbrace-\mathcal{D}(\mathsf{F}||\mathsf{F}_{\mathsf{S}})\right\rbrace}d\mathbf{f}^{}_{\mathsf{S}}d\mathbf{f}^{}_{\mathsf{T}}
	\nonumber
	\\
	&\hspace{-30pt}+
	\log c_{\mathsf{M}^{\mathsf{o}}}-\log c_{\mathsf{M}^{\mathsf{o}}}
	\nonumber
	\\
	&\hspace{-40pt}=
	\int\mathsf{M}(\mathbf{f}^{}_{\mathsf{S}},\mathbf{f}^{}_{\mathsf{T}}|\mathsf{F}_{\mathsf{S}},\mathbf{y}^{}_{\mathsf{T}},\theta)
	\log\bigg[\frac{\mathsf{M}(\mathbf{f}^{}_{\mathsf{S}},\mathbf{f}^{}_{\mathsf{T}}|\mathsf{F}_{\mathsf{S}},\mathbf{y}^{}_{\mathsf{T}},\theta)}{\mathsf{M}^{\mathsf{o}}(\mathbf{f}^{}_{\mathsf{S}},\mathbf{f}^{}_{\mathsf{T}}|\mathsf{F}_{\mathsf{S}},\mathbf{y}^{}_{\mathsf{T}},\theta)}\bigg]d\mathbf{f}^{}_{\mathsf{S}}d\mathbf{f}^{}_{\mathsf{T}}
	\nonumber
	\\
	&\hspace{-30pt}-
	\log c_{\mathsf{M}^{\mathsf{o}}}
	,
	\nonumber
\end{align}
where the ($\mathbf{f}^{}_{\mathsf{S}}$-conditioned) KLD from $\mathsf{F}(\mathbf{y}_{\mathsf{S}}^{\star}|\mathbf{f}^{}_{\mathsf{S}},\theta)$ to $\mathsf{F}_{\mathsf{S}}(\mathbf{y}_{\mathsf{S}}^{\star}|\mathbf{y}^{}_{\mathsf{S}})$ is given by
\begin{equation}
	\mathcal{D}(\mathsf{F}||\mathsf{F}_{\mathsf{S}})
	=
	\int \mathsf{F}(\mathbf{y}_{\mathsf{S}}^{\star}|\mathbf{f}^{}_{\mathsf{S}},\theta)\log\frac{\mathsf{F}(\mathbf{y}_{\mathsf{S}}^{\star}|\mathbf{f}^{}_{\mathsf{S}},\theta)}{\mathsf{F}_{\mathsf{S}}(\mathbf{y}_{\mathsf{S}}^{\star}|\mathbf{y}^{}_{\mathsf{S}})}d\mathbf{y}_{\mathsf{S}}^{\star}
	,
	\nonumber
\end{equation}
and the normalizing constant of $\mathsf{M}^{\mathsf{o}}(\mathbf{f}^{}_{\mathsf{S}},\mathbf{f}^{}_{\mathsf{T}}|\cdot)$ is
\begin{align}
	c_{\mathsf{M}^{\mathsf{o}}}
	&=
	\int\mathsf{F}(\mathbf{f}^{}_{\mathsf{S}},\mathbf{f}^{}_{\mathsf{T}}|\mathbf{y}^{}_{\mathsf{T}},\theta)\exp\left\lbrace-\mathcal{D}(\mathsf{F}||\mathsf{F}_{\mathsf{S}})\right\rbrace d\mathbf{f}^{}_{\mathsf{S}}d\mathbf{f}^{}_{\mathsf{T}}
	.
	\nonumber
\end{align}
These rearrangements reveal the marginal distribution \eqref{fpd_optimal_posterior_distribution_generic} of the FPD-optimal model \eqref{fpd_optimal_model}.\hfill\qedsymbol

\subsection{Proof of Proposition \ref{joint_posterior_model_specific}}\label{joint_posterior_model_specific_proof}
To simplify the subsequent rearrangements, let us rewrite \eqref{fpd_optimal_posterior_distribution_generic} as follows:
\begin{align}
	\mathsf{M}^{\mathsf{o}}(\mathbf{f}^{}_{\mathsf{S}},\mathbf{f}^{}_{\mathsf{T}}|\mathsf{F}_{\mathsf{S}},\mathbf{y}^{}_{\mathsf{T}},\theta)
	&\propto
	\nonumber
	\\
	&\hspace{-24pt}\mathsf{F}(\mathbf{y}^{}_{\mathsf{T}}|\mathbf{f}^{}_{\mathsf{T}},\theta)\mathsf{F}(\mathbf{f}^{}_{\mathsf{S}},\mathbf{f}^{}_{\mathsf{T}}|\theta)\exp\left\lbrace-\mathcal{D}(\mathsf{F}||\mathsf{F}_{\mathsf{S}})\right\rbrace
	.
	\label{fpd_optimal_posterior_distribution_generic_rewritten_a}
\end{align}
After adopting \eqref{source_output_data_model} and \eqref{posterior_predictor}, the exponential term in \eqref{fpd_optimal_posterior_distribution_generic_rewritten_a} results in
\begin{equation}
	\exp\left\lbrace-\mathcal{D}(\mathsf{F}||\mathsf{F}_{\mathsf{S}})\right\rbrace
	\propto
	\mathcal{N}(\bar{\mathbf{m}}_{\mathsf{S}};\mathbf{f}_{\mathsf{S}},\bar{\mathbf{R}}_{\mathsf{S}})
	.
	\label{kld_instantiated}
\end{equation}
Substituting \eqref{target_output_data_model}, \eqref{pre_prior}, and \eqref{kld_instantiated} into \eqref{fpd_optimal_posterior_distribution_generic_rewritten_a}, we obtain
\begin{align}
	\mathsf{M}^{\mathsf{o}}(\mathbf{f}^{}_{\mathsf{S}},\mathbf{f}^{}_{\mathsf{T}}|\mathsf{F}_{\mathsf{S}},\mathbf{y}^{}_{\mathsf{T}},\theta)
	&\propto
	\nonumber
	\\
	&\hspace{-24pt}
	\mathcal{N}(\mathbf{y}_{\mathsf{T}};\mathbf{f}_{\mathsf{T}},\sigma^{2}_{\mathsf{T}}I_{n_{\mathsf{T}}})\mathcal{N}(\bar{\mathbf{m}}_{\mathsf{S}};\mathbf{f}_{\mathsf{S}},\bar{\mathbf{R}}_{\mathsf{S}})
	\nonumber
	\\
	&\hspace{18pt}
	\mathcal{N}
	\left(
	\left[
		\begin{matrix}
			\mathbf{f}_{\mathsf{S}}
			\\
			\mathbf{f}_{\mathsf{T}}
		\end{matrix}
	\right]
	\!;\!
	\left[
		\begin{matrix}
			\mathbf{0}
			\\
			\mathbf{0}
		\end{matrix}
	\right]
	\!,\!
	\left[
		\setlength\arraycolsep{2pt}
		\begin{matrix}
			\mathbf{K}_{\mathsf{S}\mathsf{S}} & \mathbf{K}_{\mathsf{S}\mathsf{T}}
			\\
			\mathbf{K}_{\mathsf{T}\mathsf{S}} & \mathbf{K}_{\mathsf{T}\mathsf{T}}
		\end{matrix}
		\right]
	\right)
	.
	\label{fpd_optimal_posterior_distribution_generic_rewritten_b}
\end{align}
To use Lemma \ref{gaussian_joint_density} with \eqref{fpd_optimal_posterior_distribution_generic_rewritten_b}, we adopt the following notational assignments:
\begin{align}
	&
	\mathbf{y}
	\equiv
	\left[
		\begin{matrix}
			\bar{\mathbf{m}}^{}_{\mathsf{S}}
			\\
			\mathbf{y}^{}_{\mathsf{T}}
		\end{matrix}
	\right]
	,
	\hspace{10pt}
	\mathbf{f}
	\equiv
	\left[
		\begin{matrix}
			\mathbf{f}^{}_{\mathsf{S}}
			\\
			\mathbf{f}^{}_{\mathsf{T}}
		\end{matrix}
	\right]
	,
	\hspace{10pt}
	\mathbf{m}
	\equiv
	\left[
		\begin{matrix}
			\mathbf{0}
			\\
			\mathbf{0}
		\end{matrix}
	\right]
	,
	\nonumber
	\\[5pt]
	&
	\hspace{-10pt}
	\mathbf{\Sigma}
	\equiv
	\left[
		\setlength\arraycolsep{2pt}
		\begin{matrix}
			\bar{\mathbf{R}}^{}_{\mathsf{S}} & \mathbf{O}
			\\
			\mathbf{O} & \sigma^{2}_{\mathsf{T}}I_{n_{\mathsf{T}}}
		\end{matrix}
	\right]
	,
	\hspace{20pt}
	\mathbf{K}
	\equiv
	\left[
		\setlength\arraycolsep{2pt}
		\begin{matrix}
			\mathbf{K}_{\mathsf{S}\mathsf{S}} & \mathbf{K}_{\mathsf{S}\mathsf{T}}
			\\
			\mathbf{K}_{\mathsf{T}\mathsf{S}} & \mathbf{K}_{\mathsf{T}\mathsf{T}}
		\end{matrix}
	\right]
	,
	\label{blocks}
\end{align}
which allows us to write
\begin{align}
	\mathsf{M}^{\mathsf{o}}(\mathbf{f}^{}_{\mathsf{S}},\mathbf{f}^{}_{\mathsf{T}}|\mathsf{F}_{\mathsf{S}},\mathbf{y}^{}_{\mathsf{T}},\theta)
	&\propto
	\label{fpd_optimal_posterior_distribution_generic_rewritten_c}%\nonumber
	\\[5pt]
	&\hspace{-74pt}
	\mathcal{N}
	\left(
		\left[
			\begin{matrix}
				\mathbf{f}_{\mathsf{S}}
				\\
				\mathbf{f}_{\mathsf{T}}
				\\
				\bar{\mathbf{m}}_{\mathsf{S}}
				\\
				\mathbf{y}_{\mathsf{T}}
			\end{matrix}
		\right]
		\!;\!
		\left[
			\begin{matrix}
				\mathbf{0}
				\\
				\mathbf{0}
				\\
				\mathbf{0}
				\\
				\mathbf{0}
			\end{matrix}
		\right]
		\!,\!
		\left[
			\setlength\arraycolsep{3pt}
			\begin{matrix}
				\mathbf{K}_{\mathsf{S}\mathsf{S}} & \mathbf{K}_{\mathsf{S}\mathsf{T}} & \!\!\mathbf{K}_{\mathsf{S}\mathsf{S}} & \!\!\!\!\mathbf{K}_{\mathsf{S}\mathsf{T}}
				\\
				\mathbf{K}_{\mathsf{T}\mathsf{S}} & \mathbf{K}_{\mathsf{T}\mathsf{T}} & \!\!\mathbf{K}_{\mathsf{T}\mathsf{S}} & \!\!\!\!\mathbf{K}_{\mathsf{T}\mathsf{T}}
				\\
				\mathbf{K}_{\mathsf{S}\mathsf{S}} & \mathbf{K}_{\mathsf{S}\mathsf{T}} & \!\!\mathbf{K}_{\mathsf{S}\mathsf{S}} + \bar{\mathbf{R}}_{\mathsf{S}} & \!\!\!\!\mathbf{K}_{\mathsf{S}\mathsf{T}}
				\\
				\mathbf{K}_{\mathsf{T}\mathsf{S}} & \mathbf{K}_{\mathsf{T}\mathsf{T}} & \!\!\mathbf{K}_{\mathsf{T}\mathsf{S}} & \!\!\!\!\mathbf{K}_{\mathsf{T}\mathsf{T}} + \sigma^{2}_{\mathsf{T}}I_{n_{\mathsf{T}}}
			\end{matrix}
		\right]
	\right)
	\!.
	\nonumber
\end{align}
Consequently, applying Lemma \ref{gaussian_conditional_and_marginal_densities} to \eqref{fpd_optimal_posterior_distribution_generic_rewritten_c}---with the same blocks as in \eqref{blocks}---we obtain
\begin{equation}
	\mathsf{M}^{\mathsf{o}}(\mathbf{f}^{}_{\mathsf{S}},\mathbf{f}^{}_{\mathsf{T}}|\mathsf{F}_{\mathsf{S}},\mathbf{y}^{}_{\mathsf{T}},\theta)
	=
	\mathcal{N}(\mathbf{m}^{\mathsf{o}},\mathbf{K}^{\mathsf{o}})
	,
	\nonumber
\end{equation}
where
\begin{align}
	\mathbf{m}^{\mathsf{o}}
	&\equiv
	\left[
		\begin{matrix}
			\mathbf{m}^{\mathsf{o}}_{\mathsf{S}}
			\\
			\mathbf{m}^{\mathsf{o}}_{\mathsf{T}}
		\end{matrix}
	\right]
	\equiv
	\left[
		\begin{matrix}
			m^{\mathsf{o}}_{\mathsf{S}}(\mathbf{x}_{\mathsf{S}})
			\\
			m^{\mathsf{o}}_{\mathsf{T}}(\mathbf{x}_{\mathsf{T}})
		\end{matrix}
	\right]
	,
	\nonumber
	\\
	\mathbf{K}^{\mathsf{o}}
	&\equiv
	\left[
		\setlength\arraycolsep{2pt}
		\begin{matrix}
			\mathbf{K}^{\mathsf{o}}_{\mathsf{S}\mathsf{S}} & \mathbf{K}^{\mathsf{o}}_{\mathsf{S}\mathsf{T}}
			\\
			\mathbf{K}^{\mathsf{o}}_{\mathsf{T}\mathsf{S}} & \mathbf{K}^{\mathsf{o}}_{\mathsf{T}\mathsf{T}}
		\end{matrix}
	\right]
	\equiv
	\left[
		\setlength\arraycolsep{2pt}
		\begin{matrix}
			k^{\mathsf{o}}_{\mathsf{S}\mathsf{S}}(\mathbf{x}_{\mathsf{S}},\mathbf{x}_{\mathsf{S}}) & k^{\mathsf{o}}_{\mathsf{S}\mathsf{T}}(\mathbf{x}_{\mathsf{S}},\mathbf{x}_{\mathsf{T}})
			\\
			k^{\mathsf{o}}_{\mathsf{T}\mathsf{S}}(\mathbf{x}_{\mathsf{T}},\mathbf{x}_{\mathsf{S}}) & k^{\mathsf{o}}_{\mathsf{T}\mathsf{T}}(\mathbf{x}_{\mathsf{T}},\mathbf{x}_{\mathsf{T}})
		\end{matrix}
	\right]
	,
	\nonumber
\end{align}
which is subblock-wise computed, $q,p\in(\mathsf{S},\mathsf{T})$, as
\begin{align}
	m_{q}^{\mathsf{o}}(\mathbf{x}_{q})
	&=
	\mathbf{k}_{q}(\mathbf{K}+\operatorname{blkdiag}(\bar{\mathbf{R}}_{\mathsf{S}},\sigma^{2}_{\mathsf{T}}I_{n}))^{-1}
	\left[
		\begin{matrix}
			\bar{\mathbf{m}}_{\mathsf{S}}
			\\
			\mathbf{y}_{\mathsf{T}}
		\end{matrix}
	\right]
	,
	\nonumber
	\\
	k_{qp}^{\mathsf{o}}(\mathbf{x}_{q},\mathbf{x}_{p})
	&=
	k_{qp}(\mathbf{x}_{q},\mathbf{x}_{p})
	\nonumber
	\\
	&\hspace{18pt}-\mathbf{k}_{q}(\mathbf{K}+\operatorname{blkdiag}(\bar{\mathbf{R}}_{\mathsf{S}},\sigma^{2}_{\mathsf{T}}I_{n}))^{-1}\mathbf{k}_{p}
	,
	\nonumber
\end{align}
and
\begin{equation}
	\mathbf{k}_{q}
	\equiv
	\left[
		\setlength\arraycolsep{2pt}
		\begin{matrix}
			k_{q\mathsf{S}}(\mathbf{x}_{q},\mathbf{x}_{\mathsf{S}}) & k_{q\mathsf{T}}(\mathbf{x}_{q},\mathbf{x}_{\mathsf{T}})
		\end{matrix}
	\right]
	,
	\hspace{20pt}
	\mathbf{k}_{p}
	\equiv
	\left[
		\begin{matrix}
			k_{\mathsf{S}p}(\mathbf{x}_{\mathsf{S}},\mathbf{x}_{p}) \\ k_{\mathsf{T}p}(\mathbf{x}_{\mathsf{T}},\mathbf{x}_{p})
		\end{matrix}
	\right]
	.
	\nonumber
\end{equation}
Now, extending both $\mathbf{x}_{q}$ and $\mathbf{x}_{p}$ with $x$ leads to
\begin{equation}
	\mathsf{M}^{\mathsf{o}}(\mathbf{f}^{}_{\mathsf{S}},\mathbf{f}^{}_{\mathsf{T}},f^{}_{\mathsf{S}},f^{}_{\mathsf{T}}|\mathsf{F}_{\mathsf{S}},\mathbf{y}^{}_{\mathsf{T}},\theta,x)
	=
	\mathcal{N}(\mathbf{m}_{x}^{\mathsf{o}},\mathbf{K}_{x}^{\mathsf{o}})
	,
	\nonumber
\end{equation}
where
\begin{align}
	\mathbf{m}_{x}^{\mathsf{o}}
	&\equiv
	\left[
		\begin{matrix}
			m^{\mathsf{o}}_{\mathsf{S}}(\mathbf{x}_{\mathsf{S}})
			\\
			m^{\mathsf{o}}_{\mathsf{T}}(\mathbf{x}_{\mathsf{T}})
			\\
			m^{\mathsf{o}}_{\mathsf{S}}(x)
			\\
			m^{\mathsf{o}}_{\mathsf{T}}(x)
		\end{matrix}
	\right]
	,
	\nonumber
	\\[5pt]
	\mathbf{K}_{x}^{\mathsf{o}}
	&\equiv
	\left[
		\setlength\arraycolsep{2pt}
		\begin{matrix}
			k^{\mathsf{o}}_{\mathsf{S}\mathsf{S}}(\mathbf{x}_{\mathsf{S}},\mathbf{x}_{\mathsf{S}}) & k^{\mathsf{o}}_{\mathsf{S}\mathsf{T}}(\mathbf{x}_{\mathsf{S}},\mathbf{x}_{\mathsf{T}}) & k^{\mathsf{o}}_{\mathsf{S}\mathsf{S}}(\mathbf{x}_{\mathsf{S}},x') & k^{\mathsf{o}}_{\mathsf{S}\mathsf{T}}(\mathbf{x}_{\mathsf{S}},x')
			\\
			k^{\mathsf{o}}_{\mathsf{T}\mathsf{S}}(\mathbf{x}_{\mathsf{T}},\mathbf{x}_{\mathsf{S}}) & k^{\mathsf{o}}_{\mathsf{T}\mathsf{T}}(\mathbf{x}_{\mathsf{T}},\mathbf{x}_{\mathsf{T}}) & k^{\mathsf{o}}_{\mathsf{T}\mathsf{S}}(\mathbf{x}_{\mathsf{T}},x') & k^{\mathsf{o}}_{\mathsf{T}\mathsf{T}}(\mathbf{x}_{\mathsf{T}},x')
			\\
			k^{\mathsf{o}}_{\mathsf{S}\mathsf{S}}(x,\mathbf{x}_{\mathsf{S}}) & k^{\mathsf{o}}_{\mathsf{S}\mathsf{T}}(x,\mathbf{x}_{\mathsf{T}}) & k^{\mathsf{o}}_{\mathsf{S}\mathsf{S}}(x,x') & k^{\mathsf{o}}_{\mathsf{S}\mathsf{T}}(x,x')
			\\
			k^{\mathsf{o}}_{\mathsf{T}\mathsf{S}}(x,\mathbf{x}_{\mathsf{S}}) & k^{\mathsf{o}}_{\mathsf{T}\mathsf{T}}(x,\mathbf{x}_{\mathsf{T}}) & k^{\mathsf{o}}_{\mathsf{T}\mathsf{S}}(x,x') & k^{\mathsf{o}}_{\mathsf{T}\mathsf{T}}(x,x')
		\end{matrix}
	\right]
	,
	\nonumber
\end{align}
which---after marginalizing w.r.t.\ $(\mathbf{f}_{\mathsf{S}},\mathbf{f}_{\mathsf{T}})$---yields the sought result \eqref{fpd_optimal_posterior_distribution_specific}.\hfill\qedsymbol

\balance
\bibliographystyle{IEEEtran}
\bibliography{IEEEabrv,main}

\end{document}